\documentclass[journal]{IEEEtran}
\usepackage{graphicx}
\usepackage{multirow}
\usepackage{caption}
\usepackage{supertabular}
\usepackage{lineno,hyperref}
\modulolinenumbers[5]
\usepackage{makecell}
\usepackage[square, comma, sort&compress, numbers]{natbib}
\usepackage[labelformat=simple]{subcaption}

\usepackage{color}
\usepackage{amsfonts}
\usepackage{algorithm}
\usepackage{algpseudocode}

\usepackage{latexsym,bm}
\usepackage{amsmath,amssymb}
\usepackage{colortbl}

\ifCLASSINFOpdf
\else
\fi

\begin{document}
%
\title{Multimodal Gait Recognition for Neurodegenerative Diseases}

\author{
Aite Zhao, Jianbo Li, Junyu Dong, Lin Qi, Qianni Zhang, Ning Li, Xin Wang and Huiyu Zhou
\thanks{Aite Zhao and Jianbo Li are with College of Computer Science and Technology, Qingdao University, China. E-mail: zhaoaite@qdu.edu.cn; lijianbo@qdu.edu.cn.}
\thanks{Junyu Dong and Lin Qi are with College of Information Science and Engineering, Ocean University of China, China. E-mail: dongjunyu@ouc.edu.cn; qilin@ouc.edu.cn}
\thanks{Qianni Zhang is with School of Electronic Engineering and Computer Science, Queen Mary University of London, UK. E-mail: qianni.zhang@qmul.ac.uk.}
\thanks{Ning Li is with College Of Electronic and Information Engineering, Nanjing University of Aeronautics and Astronautics, China. E-mail: lnuaa@me.com.}
\thanks{Xin Wang is with College of Computer and Information, Hohai University, China. E-mail: wang\_xin@hhu.edu.cn.}
\thanks{Huiyu Zhou is with School of Informatics, University of Leicester, Leicester, LE1 7RH, U.K. E-mail: hz143@leicester.ac.uk.}
\thanks{(Corresponding authors: Aite Zhao and Huiyu Zhou.)}
\vspace{-1cm}}

\maketitle

\begin{abstract}
In recent years, single modality based gait recognition has been extensively explored in the analysis of medical images or other sensory data, and it is recognised that each of the established approaches has different strengths and weaknesses. As an important motor symptom, gait disturbance is usually used for diagnosis and evaluation of diseases; moreover, the use of multi-modality analysis of the patient's walking pattern compensates for the one-sidedness of single modality gait recognition methods that only learn gait changes in a single measurement dimension. Fusion of multiple measurement resources has demonstrated promising performance in the identification of gait patterns associated with individual diseases. In this paper, as an useful tool, we propose a novel hybrid model to learn the gait differences between three neurodegenerative diseases, between patients with different severity levels of Parkinson's disease and between healthy individuals and patients, by fusing and aggregating data from multiple sensors. A spacial feature extractor (SFE) is applied to generating representative features of images or signals. In order to capture temporal information from the two modality data, a new correlative memory neural network (CorrMNN) architecture is designed for extracting temporal features. Afterwards, we embed a multi-switch discriminator to associate the observations with individual state estimations. Compared with several state of the art techniques, our proposed framework shows more accurate classification results.


\end{abstract}

\IEEEpeerreviewmaketitle

\section{Introduction}
\IEEEPARstart{N}{eurodegenerative} diseases (NDDs) are described as disorders with selective loss of neurons and distinct involvement of functional systems. Since flexion and extension motions of two lower limbs are regulated by the central nervous system, the gait of a patient with a neurodegenerative disorder would become abnormal due to deterioration of motor neurons. Thus, analysis of gait parameters is invaluable for better understanding of the mechanisms of movement disorders and the development of NDDs. Patients with neurodegenerative diseases have their own unique gait patterns \cite{Ren2016Gait}. For example, Parkinson's disease (PD) and Huntington’s disease (HD) are typical disorders of the basal ganglia and are associated with characteristic changes in gait rhythms, including those from unstable, festinating, short and freezing gaits \cite{Koller1985The,Ahn2017Smart}. Amyotrophic lateral sclerosis (ALS) is a disorder, primarily affecting the motoneurons of the cerebral cortex, brain stem, and spinal cord, leading to scissors and spastic gaits \cite{Kandel2000Principles}. Moreover, gaits of patients with different severity degrees of NDDs are of various appearances that can be quantified by the unified Parkinson’s disease rating scale (UPDRS), unified Huntington's disease rating scale (UHDRS), and Hoehn \& Yahr rating scale (H \& Y) \cite{ZHAO20181,Bai2017Quantification,Koller1985The,Vasquez2018Multimodal}. In this paper, we design a learning framework to effectively analyse the gait characteristics and identify individual diseases and disease severity.


\begin{figure}
  \centering
  \includegraphics[width=8.5cm,height=5cm]{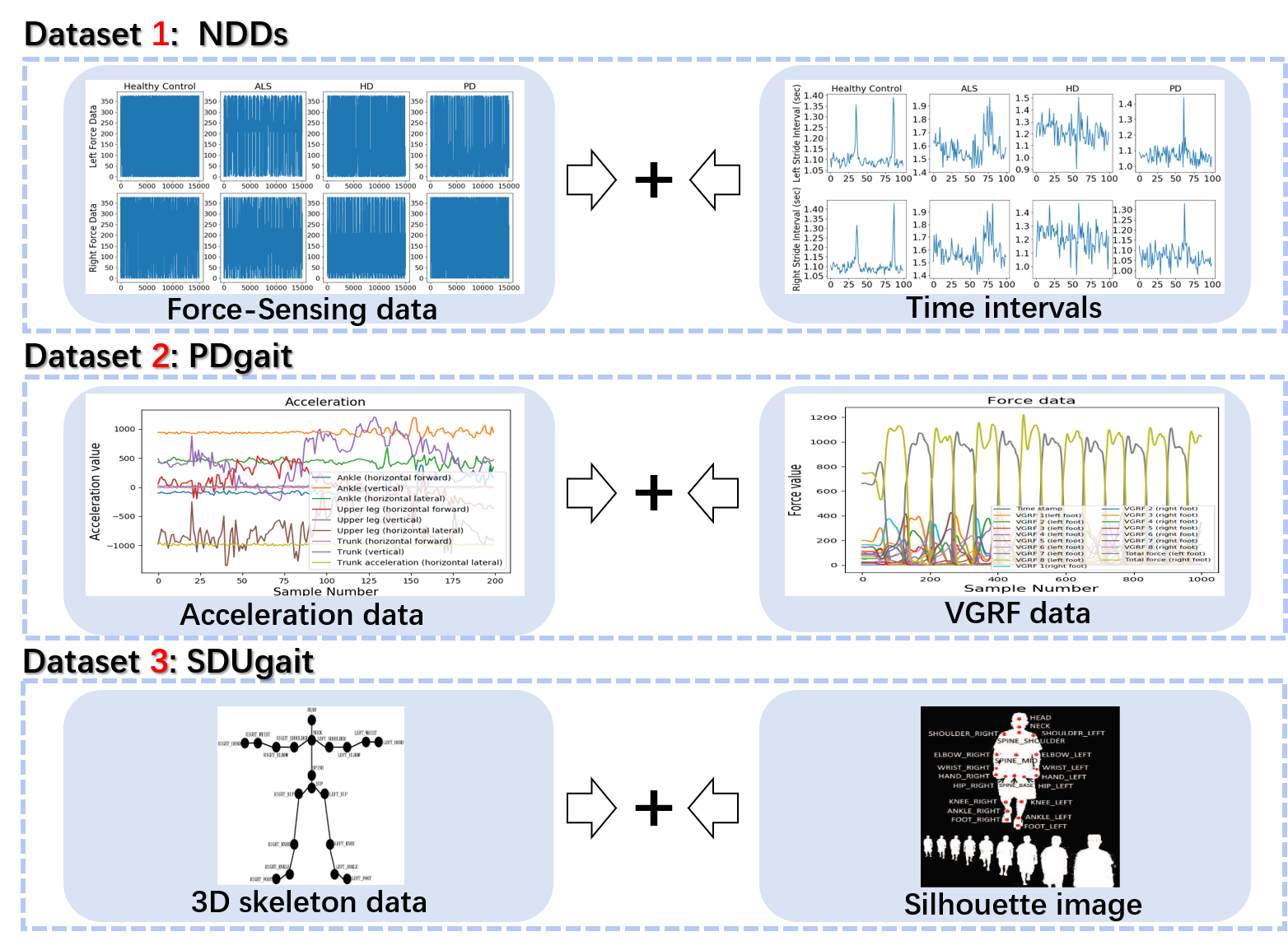}
  \caption{The multimodal gait data used in this project: We evaluate the proposed approach over three datasets, namely NDDs, PDgait and SDUgait, each of which consists of bimodal data, i.e., force-sensing and time intervals, acceleration and VGRF data, 3D skeleton data and silhouette images, respectively.}
   \label{data}
   \vspace{-0.5cm}
\end{figure}

A large amount of studies are based on automated gait analysis using the data collected by different sensors, such as Kinect, cameras, accelerators, or force-sensing sensors \cite{7888946,7452580,Zeng2015Classification,7938715}. Previous studies have reported that using one of these data resources results in promising detection and diagnosis of NDDs by gait recognition. For example, using time series data gathered by chronoscopes, \cite{Zeng2015Classification} reports the identification of healthy controls and patients with neurodegenerative diseases via deterministic learning. However, gait recognition requires rich fine-grained features, as differences between different gait styles are usually much more subtle than those between common action categories. To capture the subtle features without losing other key biomarkers, we use different sensors to collect data that describe multi-modality gait characteristics. To obtain more precise recognition results, the gait data collected by diverse sensors need to be jointly exploited rather than individually processed. An effective multi-modal fusion approach is proposed here, which plays the critical role of fusing data in multiple modalities and generating joint prediction of gait recognition.



Several multi-modal approaches exploit the co-occurrence relationship between audio, image and video acquired by different sensors. For example, research studies \cite{Filippo2016Human,Poria2018MELD} have considered the use of heterogeneous data collected from accelerometers, smartphone or cameras for recognition tasks, which demonstrated significant system improvement over single-modality approaches. \cite{Poria2018MELD} adopts convolutional neural networks (CNNs) and long short-term memory (LSTM) to analyse audio, visual, and textual data in the applications such as conversation, and shows the impact of contextual and multi-modal information on emotion recognition during conversation. It does not only extract specific spacial features of facial expression instantly, but also capture the temporal changes of human emotion. However, very few works investigate the inherent correlation of different data resources, and use this correlation to recognise the micro-variations of NDDs.


Inspired by recent progress in machine learning, a novel hybrid gait recognition network is proposed here to discover relevant and decisive attributes of the input signals or images, which helps extract and integrate spatial and temporal descriptors selectively. Without loss of generality, in this paper, we fuse two different data sources to produce comprehensive knowledge that includes not only data from different views but also, in some cases, unique foot movement symptoms that would be otherwise inaccessible to describe gait characteristics of patients with NDDs being monitored. In the description of gait, we select two types of gait data, including time intervals (in seconds), force-sensing data (in Newton), vertical ground reaction force (VGRF in Newton), acceleration, images and 3D skeleton data (Fig. \ref{data}). We will focus on the features of the gait data and explain their values in the following sections.

As mentioned in \cite{7835604}, power distributions over feet vary between NDDs and healthy subjects. Force-sensing data measures the variations of forces whilst stride time intervals illustrate the micro changes of gait. Gait images have an overall representation of human body structure, and acceleration and 3D skeleton data focus on the micro-motion of joints and the change of gait speeds. Gait cycle fluctuations during the steady-state of human walking, and variations in a series of stride intervals between time instants of consecutive ground-contact events for either foot exhibit subtle changes. The challenges of our research is to learn and fuse these dynamic changes.


However, fusing multiple resources is not trivial. For instance, multiple sensor data differ greatly in forms and scales. In our system, we design a spacial feature extractor (SFE) and a correlative memory neural network (CorrMNN) scheme to extract spatio-temporal features, and then use a multi-switch discriminator to establish a diagnostic model which determines whether the input gait samples come from NDDs or normal people.


On the one hand, to highlight unique spacial characteristics of the ordinal data, we use SFE to undertake feature extraction and fusion concurrently. It has built-in Fisher vector generators to map the overall bimodal data onto the same feature space, encode and aggregate the local features by sum pooling over the input data. Linear Discriminant Analysis (LDA) is used to reduce the data dimensionality to obtain a salient yet light spacial feature vector. On the other hand, CorrMNN components include an interconnected dual channel memory neural network (DCMNN) with a correlation computing unit, which focuses on the temporal and contextual representations and computes the correlation between the two types of data. During the training process of CorrMNN, we use an unsupervised method to maximize the correlation of features in the same class whilst using a supervised method to maximize the differences of features in different classes to achieve accurate classification. Afterwards, using the spatial-temporal features of gait, we expect to construct a multi-switch determinator for NDDs and disease severity classification due to its strong modeling and time-sensitive abilities. Here, hidden Markov models (HMMs) are employed to simulate the states with properly defined parameters to distinguish the classes.

The main contributions of our work are summarized as follows:

\begin{itemize}
	\item A spacial feature extractor (SFE) is constructed to learn the spatial information of multi-frame time series data and output spacial features after dimensionality reduction.
	\item A correlative memory neural network (CorrMNN) is designed to measure the correlation in bimodal gait data and extract the dynamic changes of gait to generate temporal features. In spite of its use in analysing bimodal gait data herewith, the proposed CorrMNN can be easily adapted to deal with the data of over two modalities. 
    \item A multi-switch discriminator is built for classifying the input gait patterns and rating the disease severity.
    \item The proposed hybrid model can selectively integrate multi-modal gait features as well as extract critical spatio-temporal information of gait for identifying neurodegenerative diseases and disease severity.
    \item The proposed method is comprehensively evaluated over two challenging neurodegenerative datasets and one normal gait dataset for multimodal gait recognition.
\end{itemize}

The state of the art technologies compared in this paper include single- and multi-modal approaches. The single-modal methods are:
\begin{itemize}
\setlength{\itemsep}{0pt}
\setlength{\parsep}{0pt}
\setlength{\parskip}{0pt}
\item LSTM (long-short term memory) is a model with expandable nodes, suitable for temporal data.
\item BiLSTM (bi-directional long-short term memory) is a combination of forward and backward LSTMs.
\item GRU (gate recurrent unit) is a lightweight variant of LSTM with fewer variables.
\item CNN (convolutional neural network) is a kind of feed-forward neural network with a deep structure and convolution calculation.
\item CapsNet (capsule network) \cite{Sabour2017Dynamic} is a neural network sensitive to spatial structures and relative positions, an upgraded version of CNN.
\item Hand-crafted Method \cite{7532940} computes static and dynamic features of gaits.
\item LSTM+CNN \cite{ZHAO20181} is a fusion model of LSTM and CNN.
\item Meta-classifiers \cite{S2014Gait} fuse random trees and bagging classifiers for NDDs classification.
\item HMM (hidden Markov model) \cite{khorasani2014hmm} is a network with a chain structure and hidden states, suitable for time series.
\item QBC (quadratic Bayes classifier) \cite{Banaie2011Introduction} is a novel classifier based on the standard Bayes classifier.
\item RBF+DL \cite{Zeng2015Classification} combines a radial basis function (RBF) neural network with deterministic learning for NDDs classification.
\item Q-BTDNN \cite{jane2016q} is designed for identifying statistic and dynamic gait information through a deep convolutional neural network.
\item C-FuzzyEn+SVM \cite{Yi2016Symmetry} calculates cross-fuzzy entropy for gait symmetry analysis and uses support vector machine (SVM) for classification.
\item PE+SVM \cite{Xia2016A} uses permutation entropy and SVM for NDDs classification.
\item 3D-STCNN \cite{HUYNHTHE2020} is designed for identifying statistic and dynamic gait information through a deep convolutional neural network
\end{itemize}

Multi-modal comparison methods consist of:
\begin{itemize}
\setlength{\itemsep}{0pt}
\setlength{\parsep}{0pt}
\setlength{\parskip}{0pt}
\item DCLSTM \cite{Zhao2018Dual} is a dual channel LSTM designed for the fusion of two data sources.
\item DCCA (deep canonical correlation analysis) extracts features by calculating the correlation between two sets of data based on a deep neural network.
\item KCCA (kernel canonical correlation analysis) extracts features by calculating the correlation between two sets of data based on the kernel function.
\item MELD-LSTMs \cite{Poria2018MELD} adopts CNN and LSTM to analyse audio, visual, and textual data.
\item RC-LSTMs (reservoir computing LSTMs) \cite{Filippo2016Human} exploits the strengths of the reservoir computing approach for modeling RNNs, appropriate for the identification and classification of temporal patterns in time series.
\end{itemize}

The rest of this paper is organized as follows. Related work is reviewed in Section II. Section III introduces the proposed framework and the details of each module. Section IV provides the results of the system evaluation. Conclusion is given in Sections V. The source code of our proposed framework will be published at https://github.com/zhaoaite/CorrMNN.

\section{Related Work}
We here explore state-of-the-art technologies in the following two areas of interests: Single-modal and multi-modal gait recognition.

\textbf{a) Single-modal gait recognition:} Heterogeneous data can be obtained from different sensors including force, speed, vision, inertial sensors and timer, which allow us to extract features in different dimensions and formats. Single-modal gait recognition based on any of these sensory readings has been a popular research topic in the last 20 years.

With the advent of cheap and easy-to-use depth sensors, such as Kinect, 3D skeleton-based gait recognition has been successfully used for diagnosis, treatment and rehabilitation of NDDs. For example, Athina \textit{et al.} presented a two-layer LSTM to analyze the movement patterns of PD patients in order to detect statistically significant differences between two impairment levels based on motor scores and performance in the unified Parkinson's disease rating scale (UPDRS) \cite{Grammatikopoulou}. Dawn \textit{et al.} used separate multi-variable linear regression models and proportional odds regression models to evaluate the associations of various Kinect variables with the UPDRS-motor outcomes by analyzing the gait of patients with Parkinson's disease \cite{Tan2019}. Lacramioara \textit{et al.} built a Kinect-based system that can distinguish between two PD stages using different classification methods (e.g. decision trees, Bayesian networks, neural networks and K-nearest neighbours classifiers), in an attempt to identify the most suitable classification method \cite{Dranca2018}. The skeletal data includes 3D coordinates of limbs but does not supply step intervals or swing duration, which are very important to describe gaits. For example, slow foot lifting and walking interval are the key descriptors to discriminate between normal and abnormal gaits.

To handle the time series data, a number of methods have been developed for extracting temporal information from data. For example, Zeng \textit{et al.} fused a radial basis function (RBF) neural network and deterministic learning to separate the time sequences of gaits of healthy controls and patients with three NDDs \cite{Zeng2015Classification}. Mikos \textit{et al.} provided patients certain adaptivity in real-time through semi-supervised neural networks using the stream of the unlabeled temporal data generated for the detection of freezing gaits \cite{Mikos2018Real}. On the other hand, generalized linear regression analysis (GLRA) and support vector machine (SVM) techniques have been deployed to implement nonlinear gait pattern classification in PD \cite{Wu2017Measuring}, where the temporal difference in the gait cycle and the ground reflection forces of feet can be fully exploited.

To deal with the gait data gathered by force and inertial sensors, which can sense subtle changes in the lower extremities and are mainly used in experimental or real medical scenes, several machine learning approaches have been designed for NDDs gait recognition. Jane \textit{et al.} designed a Q-backpropagated time delay neural network (Q-BTDNN) for severity prediction of PD using vertical ground reaction force (VGRF, in Newton) of feet \cite{jane2016q}. Another fusion model developed by Prateek \textit{et al.} aims to detect the onset and duration of the freezing of gait (FOG) for the patients with Parkinson's disease in real time using inertial sensors \cite{Prateek2018Modeling}.

\textbf{b) Multi-modal gait recognition:} Although the methods described above have achieved promising results, they can only handle single-modality data, and lack integration of modes in various configurations. We will summarise the latest techniques for multi-modal gait recognition applications in this subsection.

3-D convolutional neural networks (CNNs) and LSTMs have been utilized to extract spatial and temporal features of gait data and inertial data \cite{PradeepMultimodal}. In spite of the use of both spatiotemporal characteristics and multi-modal information, it does not consider the correlation between multi-modal resources. For example, Valentin \textit{et al.} focused on four variants of deep neural networks that were based either on fully-connected deep neural networks or CNNs to interpret the users' gait data and other activities captured by multi-sensor systems \cite{RaduMultimodal}. In addition, a comparative study of different CNN architectures was presented based on three low-level features including gray pixels, optical flows and depth maps \cite{Castro2018Multimodal}.

However, these advanced multi-modal gait recognition methods are rarely used for NDDs diagnosis. For instance, a multi-modal assessment was presented to acquire information from speech, handwriting, and gait to model the transitions of human movement and to train a CNN to classify between PD patients and healthy subjects \cite{Vasquez2018Multimodal}. Alvarez \textit{et al.} proposed a system that can continuously monitor patients with Parkinson's and Alzheimer's diseases by analyzing multi-modal data collected by innovative imaging sensors and Internet of Things devices \cite{Alvarez2018Behavior}.

Moreover, there are also many multi-modal approaches used in different research fields, such as gesture recognition \cite{Filippo2016Human,Poria2018MELD}, activity recognition and emotion recognition. However, these approaches pay little attention to both spatio-temporal information and the correlation between data from multiple sources. In our proposed work, we exploit both spatio-temporal attributes and correlative information from two different data sources and thereby increase recognition performance, while enhancing the capability of our framework for NDDs gait recognition in multi-sensor environments.

\section{Proposed Methods}
In order to achieve successful gait recognition for NDDs patients and healthy people in a less controlled environment, we here propose a multi-modal framework for spatio-temporal feature extraction and classification. Firstly, we introduce the new spacial feature extractor (SFE). Then, we describe the architecture of the correlative memory neural network (CorrMNN) for temporal feature extraction, followed by detailed discussion on the individual components of CorrMNN. Finally, we introduce the multi-switch discriminator for learning the inherent characteristics of the joint spatio-temporal features before supplying the final diagnostic result.

\subsection{The Overall Framework}
\begin{figure*}[!htb]
  \centering
  \includegraphics[width=14cm]{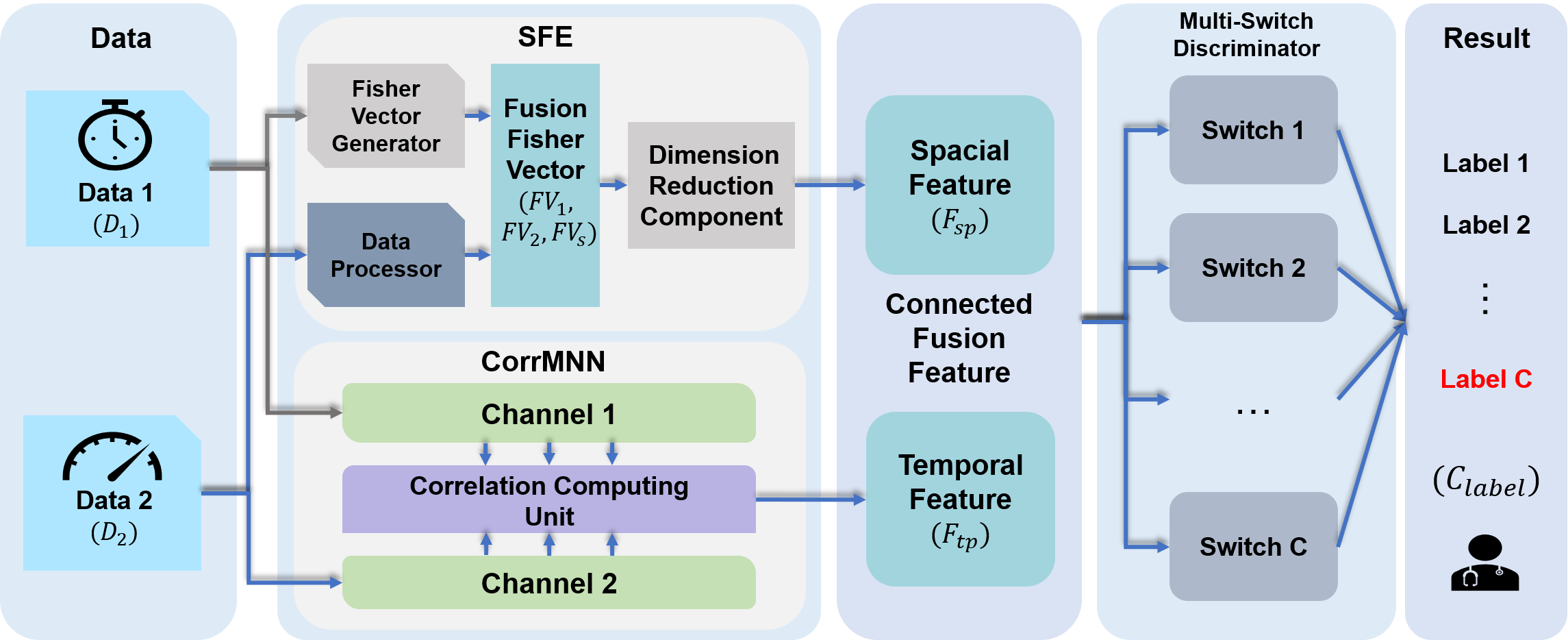}
  \caption{The structure of the proposed framework. It includes three modules: SFE for spacial feature extraction, CorrMNN for temporal feature extraction, and multi-switch discriminator for model generation and classification. Two input signals are fed into CorrMNN and SFE. The outputs are fused together and then fed into HMM switches in order to produce scores for classification.}
   \label{main_framework}
   \vspace{-0.5cm}
\end{figure*}

Our proposed framework is shown in Fig. \ref{main_framework}. In this system, two inputs from different data sources are fed into SFE and CorrMNN modules. SFE consists of a Fisher vector generator (direct generator) and a complex data processor (time- and frequency-domain generators) to handle the two inputs separately for Fisher vector generation. Afterwards, Linear Discriminant Analysis (LDA) is applied to reducing the dimensionality of the long Fisher vector for generating meaningful spacial features of the gait data. For the CorrMNN module, the same inputs come to a dual channel recurrent neural network for analyzing the temporal dynamic changes of the gait data, while a correlation computing unit is designed to consider the correlation between two channels for better fusion of the two gait data. After having obtained temporal features from CorrMNN, we combine the spacial and temporal features and then send them to the multi-switch discriminator that can automatically learn the patterns at each state, and also generate independent models to describe different classification outcomes. As shown in Fig. \ref{main_framework}, our system is an end-to-end discriminant framework with dual-modal inputs and diagnostic outputs to train the multi-modal gait recognition network.

First, we formulate the problem for multi-modal gait recognition. The two gait sequences are defined as $D_1=\{x_i\in\mathbb{R}^P, i=1,2,...,M\}$ and $D_2=\{z_i\in\mathbb{R}^Q, i=1,2,...,N\}$ with corresponding $C$-class label sequences $L_1$ and $L_2$, $M$ and $N$ are the sample numbers of the two sets of data respectively, $P$ and $Q$ are the feature dimensions of each frame of the two different gait data. For the task of feature extraction, our goal is to address two problems at the same time: (1) Successfully producing spacial features $F_{sp}$ of the two data sources, and (2) analyzing the correlation between the data and obtaining temporal features $F_{tp}$. For the task of classification, multi-switch hidden Markov models (HMMs) are utilized to learn the characteristics of the data and then produce the final classification results $C_{label}$. 


\subsection{Data Fusion in Feature Level Using Spacial Feature Extractor}

\begin{figure}[!htb]
  \centering
  \includegraphics[width=8cm]{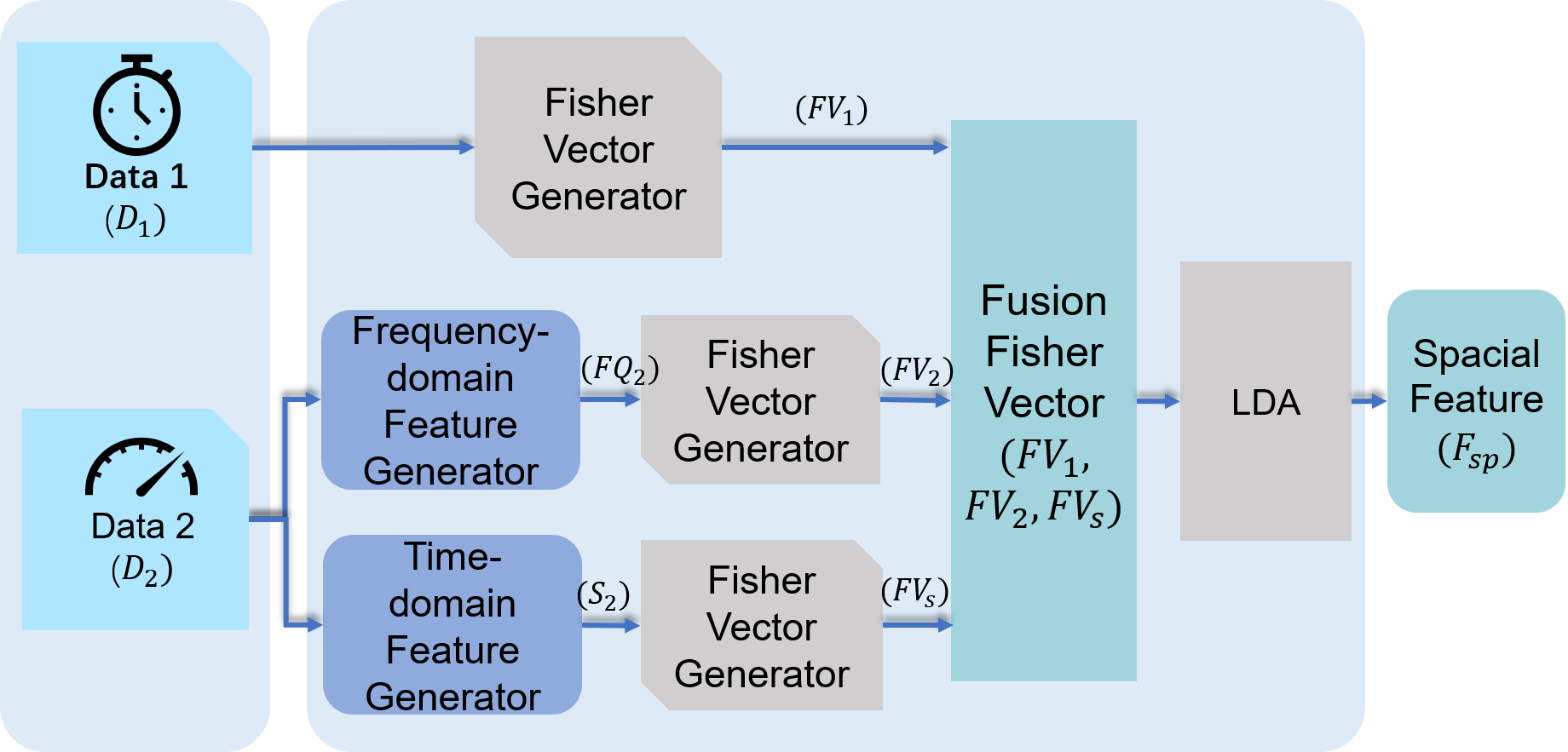}
  \caption{The SFE model. It includes three modules: Frequency- and time-domain feature extraction, Fisher vector extraction, and LDA dimensionality reduction. After several processes of the input data have been undertaken, we receive three Fisher vectors ($FV_1,FV_2,FV_s$) that represent the two-modality data, and then use LDA to derive the final 3-dimensional feature vector $F_{sp}$}.
   \label{SFE}
\end{figure}

In this section, we aim to design an efficient feature-level framework, namely SFE, to analyze multi-modal data. The corresponding bimodal data is processed separately in order to extract descriptors as the input of the Fisher vector generator. We reshape multiple frames as a training sample in advance. Since the connected Fisher vector may contain redundant information, we intend to produce more significant and effective features by dimensionality reduction. We use LDA to reduce the dimensionality of the data with the goal of minimizing the variance within the data class and maximizing the variance between the classes, resulting in better data separability.

To conduct optimal separation of data, for data type one, we first extract its descriptors and then generate the corresponding Fisher vectors. For data type two, time- and frequency-domain analysis is undertaken, leading to two observations. Formally, given a signal sequence $D_2=\{z_i\in\mathbb{R}^Q, i=1,2,...,N\}$, we employ the two-domain feature generator. The time-domain generator calculates statistics including mean, root mean square, skewness, kurtosis, waveform factor, peak factor, impulse factor and margin factor ($\Bar{X},X_{rms},S,K,W,C,I,M$) as the features of the signals ($S_2$). On the other hand, the frequency-domain generator uses discrete Fourier transform (DFT) to extract the frequency characteristics of the signals, which captures the dynamics of the data sources \cite{Yi2019An}.


After having obtained the descriptors from the two types of data, firstly, each sample of the signal $D_2=\{z_i\in\mathbb{R}^Q, i=1,2,...,N\}$ is divided into $T$ $D$-dimensional vectors $X=z_i=\{x_t \in \mathbb{R}^D,t= 1,...,T\}$, which are learned automatically. To extract the descriptors of gaits, $W_j \in \mathbb{R}^{D \times D} (j= 1,2,3)$ is defined as a description parameter matrix which can be updated according to different feature types.

We then use a GMM model (a linear combination of $K$ $D$-dimensional Gaussian distributions) to fit all the descriptors, which uses several weighted Gaussian density functions to describe the distribution of data and preserve the most frequent patterns of the original dataset by taking into account the multi-modal distribution of the numerical variables \cite{KHANMOHAMMADI2016119}. The probability of each class can be obtained from the projected samples, which can better describe the signals. Assuming that the parameter set of the Gaussian distribution includes $\lambda = \{w_k,\mu_k^d,\sigma_k^d,k=1,2,...,K\}$ (weight, mean and variance), $d={1,2,3,...,D}$ refers to the $d^{th}$ dimension of $W_jx_t$. Here, we assume that the covariance matrix is diagonal, whose diagonal elements reflect the degree of data dispersion in various dimensions. According to the standard Bayesian model, the probability that descriptor $W_jx_t$ fits the $i^{th}$ Gaussian model is $\gamma_t(i)=v(i|W_jx_t,\lambda)=w_iv_i(W_jx_t|\lambda)/(\sum_{k=1}^{K}w_kv_k(W_jx_t|\lambda))$. The probability function of GMM can be given by:
\vspace{-0.2cm}
\begin{equation}\small
\begin{aligned}
v(W_jx_t|\lambda)&=\sum_{k=1}^K w_k*v_k(W_jx_t|\lambda)\\
&=\sum_{k=1}^K {w_k} * \frac{e^{ -\frac{1}{2} {(W_jx_t-\mu_k)'} \sigma_k^{(-1)} {(W_jx_t-\mu_k)}} }  {(2 \pi) ^{(\frac{D}{2})} * {|\sigma_k|} ^ {\frac{1}{2}} }
\label{fv_v}
\end{aligned}
\end{equation}
To represent the frequency data, we denote $W_1=\{W_{ab}=(e^{-i\frac{2\pi}{D}})^{a \times b},a=1,2...,D,b=1,2...,D\}$, and use Euler's formula (${e^{xi}}= \cos{x}+i\sin{x},x\in\mathbb{R}$) to transform it to $W_1=\{W_{ab}={(\cos{\frac{2\pi}{D}}-i\sin{\frac{2\pi}{D}})}^{a \times b},a=1,2...,D,b=1,2...,D\}$ in which the real and imaginary parts are identified. Similarly, we can define $W_2\in\mathbb{R}^{D \times D}$ and $W_3\in\mathbb{R}^{D \times D}$ as the corresponding description parameter matrices to represent the data in the time-domain.

Secondly, it is assumed that each vector $W_jx_t \in \mathbb{R}^D,t= 1,...,T$ of the input signal has a distribution $p$, and when the sample number $T$ increases, the sample mean converges to the expectation $E_{x \sim p}$. In order to reduce or remove the influence of weak disturbance and noise onto the signals, we decompose the Gaussian distribution $p$ into two parts: The sample of the stronger part of the signal fluctuation has the distribution $u$. The sample of the noise of the signal fluctuation obeys the distribution $v_\lambda$. Define $0<=w<=1$ as the ratio of strong signal information against the input. We have,
\begin{equation}
p(x)=wu(x)+(1-w)v_{\lambda}(x)
\end{equation}

The likelihood function of the new hybrid distribution is ${L(X|\lambda)}=  E_{x\sim p}[\log v(W_jx_t|\lambda)]$. The new Fisher vector $G_\lambda^X$ is obtained by computing the partial derivatives of the likelihood function $L(X|\lambda)$ to $w_k$, $\mu_k^d$, $\sigma_k^d$ to represent the gradient statistics of the local descriptors.

\vspace{-0.45cm}
\begin{equation}\small
\begin{aligned}
G_{\lambda}^X&= \nabla_{\lambda} E_{x\sim p}[logv_{\lambda}(x)]\\
&=\nabla_{\lambda}\int_xp(x)logv_{\lambda}(x)dx.\\
&=w\nabla_{\lambda}\int_xu(x)logv_{\lambda}(x)dx+(1-w)\nabla_{\lambda}\int_xv_{\lambda}(x)logv_{\lambda}(x)dx
\label{fv_gradient}
\end{aligned}
\end{equation}
The parameter $\lambda$ is obtained by solving the maximum likelihood problem in the GMM modeling, that is to say, it has the form:
\vspace{-0.3cm}
\begin{equation}
\nabla_{\lambda}\int_xv_{\lambda}(x)logv_{\lambda}(x)dx=\nabla_{\lambda}E_{x\sim v_{\lambda}}[logv_{\lambda}(x)]\approx 0.
\label{fv_max0}
\end{equation}
So Eq. (\ref{fv_gradient}) is expressed as:

\begin{equation}
\begin{aligned}
G_{\lambda}^X&= w\nabla_{\lambda}\int_xu(x)logv_{\lambda}(x)dx\\
&=w\nabla_{\lambda}E_{x\sim u}[logv_{\lambda}(x)]
\end{aligned}
\label{fv_gradient1}
\end{equation}
We can see that the information of independent noisy signals is discarded in the representation of Fisher vectors, and the representation of the Fisher vectors is still related to strong signal information $w$.

Finally, the expectations of the diagonal elements of the Fisher matrix are calculated for three variables respectively to form Fishier vector ${\mathcal{G}_\lambda ^X = \{G_1,...G_{(K*(2D+1)-1))}}\} = {F_\lambda^{-1/2}}{G_\lambda^X}$
, where ${F_\lambda^{-1/2}}=E_X(G_\lambda^{X}{G_\lambda^{X}}')$ denotes the expectation while $G_\lambda^X$ denotes the partial derivatives of the corresponding variables. Afterwards, we obtain the absolute value of each element in the three normalized partial derivative components, shown in Eq. (\ref{3fvfactors}), and conduct $L2$ normalization again after having merged the vectors.
\begin{equation}\small
\left\{
\begin{array}{c}
\begin{aligned}
\mathcal{G}_{w_k}^X&=F_{w_k}^{-1/2}\frac{w  E_{x\sim u}[logv_{\lambda}(x)]}{\partial w_k}\\
\mathcal{G}_{\mu_k}^X&=F_{\mu_k}^{-1/2}\frac{w  E_{x\sim u}[logv_{\lambda}(x)]}{\partial \mu_k^d}\\
\mathcal{G}_{\sigma_k}^X&=F_{\sigma_k}^{-1/2}\frac{w  E_{x\sim u}[logv_{\lambda}(x)]}{\partial \sigma_k^d}
\label{3fvfactors}
\end{aligned}
\end{array}
\right.
\end{equation}

Assuming $\mathcal{G}_{\lambda}^{X}=\{\mathcal{G}_{w_1}^X,...,\mathcal{G}_{w_K}^X,\mathcal{G}_{\mu_1}^X,...,\mathcal{G}_{\mu_K}^X,\mathcal{G}_{\sigma_1}^X,...,\mathcal{G}_{\sigma_K}^X\}$, we normalize the Fishier vector by $L_2$ group normalization in order to emphasize the difference of three vectors in the same space, which is shown in Eq. (\ref{fv_nomalize}). We divide the three vectors, the derived gradients for different parameters, into groups to perform feature augmentation and then apply $L2$ normalization to the augmented data.

\begin{equation}\small
\begin{aligned}
&N(\mathcal{G}_{\lambda}^X)={L_2}\left(\frac{\mathcal{G}_{w_k}^X|\mathcal{G}_{w_k}^X|}{\sum_{k=1}^K|\mathcal{G}_{w_k}^X|},\frac{\mathcal{G}_{\mu_k}^X|\mathcal{G}_{\mu_k}^X|}{\sum_{k=1}^K|\mathcal{G}_{\mu_k}^X|},\frac{\mathcal{G}_{\sigma_k}^X|\mathcal{G}_{\sigma_k}^X|}{\sum_{k=1}^K|\mathcal{G}_{\sigma_k}^X|}\right)\\
\label{fv_nomalize}
\end{aligned}
\end{equation}

After having merged three Fisher vectors ($FV_1,FV_2,FV_s$), we use LDA to reduce the data dimensionality. The goal of LDA is to project a sample in a high-dimensional space onto its counterpart in a low-dimensional space so that the projection values of all the samples can hold the minimum intra-class distance and the maximum inter-class distance, resulting in better data separability. 

\subsection{CorrMNN Model}
\begin{figure}[!htb]
  \centering
  \includegraphics[width=8cm]{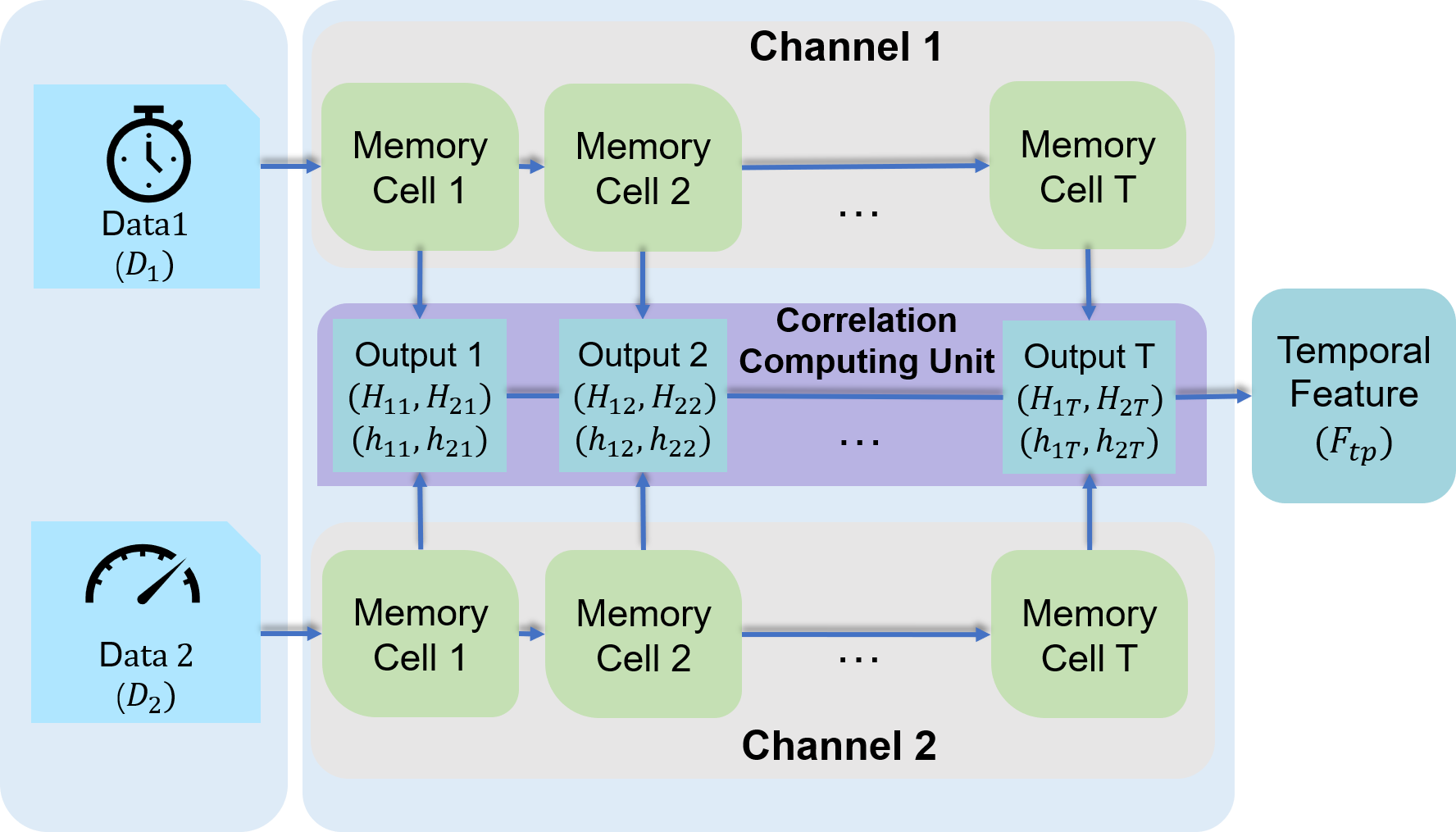}
  \caption{The structure of CorrMNN. When we train this model, each memory node of the dual channel memory neural network (DCMNN) delivers features to a unit to calculate the correlation so as to maximize the correlation between the two-channel data.}
   \label{CorrMNN}
\end{figure}

With respect to temporal feature extraction, we design a network architecture based on the architecture of long short-term memory (LSTM) \cite{Chung2014Empirical}. We apply deep canonical correlation analysis (DCCA) for generating a correlation computing unit \cite{Andrew2013Deep}, and a dual channel memory neural network (DCMNN) to build a correlative memory neural network (CorrMNN), which can process two data sources separately and then pursue correlation and classification simultaneously during the training process. 

LSTM is a gated network specially designed for time series data, which is continuously collected and well fits the two input temporal data sequences. Moreover, LSTM has the advantage of linking the observation with its contextual information and explicitly describes the characteristics at the adjacent time to reveal the internal relationship of the entire sequence. For short-length gait signals, we choose the simplest GRU of the gated control mechanism, which is suitable for processing short sequences. Besides, we only use one-layer LSTM, and three internal gated mechanism to encode the data. In addition, we add a temporary state path to ensure the integrity of gait information, making LSTM more suitable for our input data. Similarly, to handle two types of data, the dual channel memory neural network (DCMNN) bridges two data channels by computing their correlation, and fully describes their characteristics. Furthermore, we redesign the internal structure of DCCA in order to achieve better performance in CorrMNN.

The overall architecture of CorrMNN is illustrated in Fig. \ref{CorrMNN}. The input data reaches the two channels with $T$ time-expanded nodes in each channel. A deep correlation computing model is applied to receiving the output of each node in DCMNN, computing the correlation between them and minimizing the loss of distinguishing different classes in the single modality data.

\subsubsection{The Multi-Gated Memory Cell in DCMNN}
Before introducing our model, we will give a brief review of the standard LSTM. In order to model nonlinear dynamic processes, long short-term memory (LSTM) is widely used to describe temporal dynamic behaviors of time sequences \cite{hochreiter1997long}. With memory cells and three nonlinear gates (input, forget and output) in the basic LSTM structure, LSTM network is effective in learning long-term temporal dependence since these memory cells can keep their states over long time and regulate the information flows into and out of the cell.

\begin{figure}[!htb]
\centering
  \begin{center}
  \includegraphics[width=5cm]{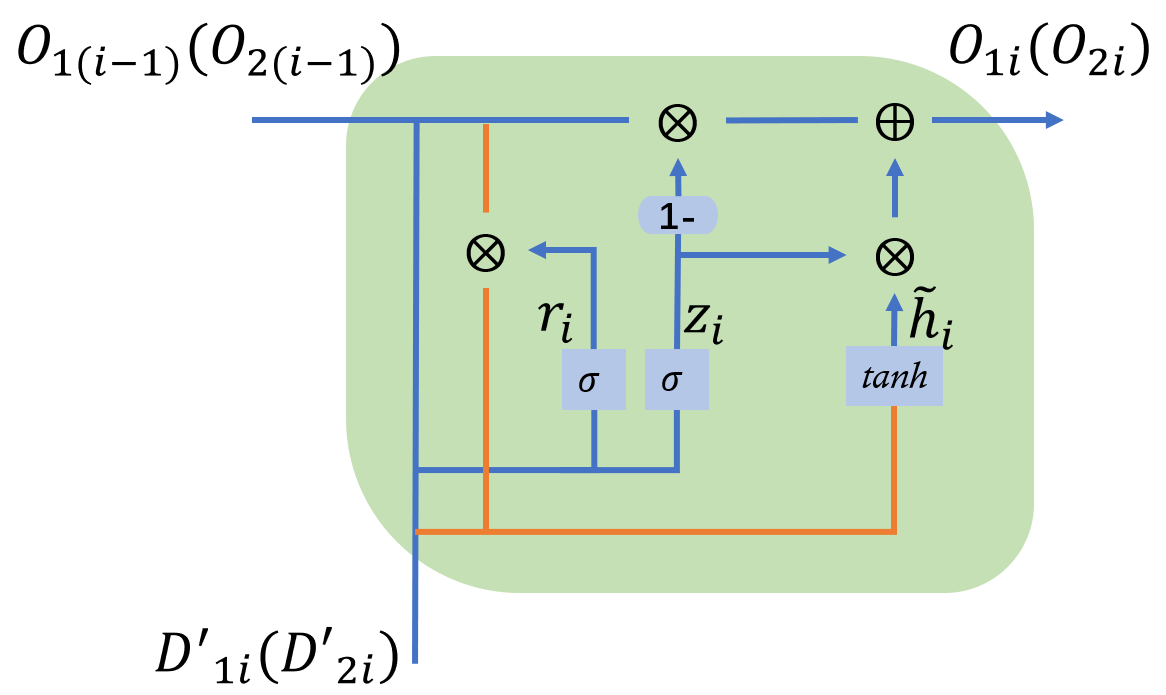}
  \caption{The original GRU cell.}
  \label{grucell}
  \vspace{-0.5cm}
  \end{center}
\end{figure}

A variant of LSTM, called the gated recurrent unit (GRU), combines the input and the forget gates with an update gate, and mixes the cell and hidden states in each node of the original LSTM, which is used as the learning model \cite{Chung2014Empirical}. GRU has fewer gates and parameters than the standard LSTM, which helps improving system efficiency to handle a large dataset. Fig. \ref{grucell} shows the structure of a GRU cell and illustrates the operations of the gates. The outputs and inputs in each cell of GRU is described in Eqs. (\ref{z_i})-(\ref{h_i}).

\begin{equation}
{z_i} = \sigma({W_z}\cdot[{O_{1(i-1)}},{D'_{1i}}])
\label{z_i}
\end{equation}
 \vspace{-0.5cm}
\begin{equation}
{r_i} = \sigma({W_r}\cdot[{O_{1(i-1)}},{D'_{1i}}])
\label{r_i}
\end{equation}
 \vspace{-0.5cm}
\begin{equation}
{{\tilde h}_i} = \tanh (W\cdot[{r_i}\otimes{O_{1(i-1)}},{D'_{1i}}])
\label{h'_i}
\end{equation}
 \vspace{-0.5cm}
\begin{equation}
{O_{1i}} = (1-{z_i})\otimes{O_{1(i-1)}}+{z_i} \otimes {{\tilde h}_i}
\label{h_i}
\end{equation}
where $D'_{1i}$ and $O_{1(i-1)}$ are the current and previous hidden states respectively in the channel 1. $\sigma$ and $\tanh$ are the logistic functions. $z_i$ indicates the output of the update gate at time step $i\in\{1,2,3...,T\}$. $r_i$ denotes the reset gate and the actual activation of the proposed unit $O_{1i}$ are then computed using Eq. (\ref{h_i}).The update gate $z$ determines whether or not the hidden state will be updated with hidden state $\tilde h$. The reset gate $r$ determines whether or not the previous hidden state is ignored. The internal parameters of the GRU nodes in channel 2 are calculated in the same way.

\begin{figure}[!htb]
  \centering
  \includegraphics[width=6cm]{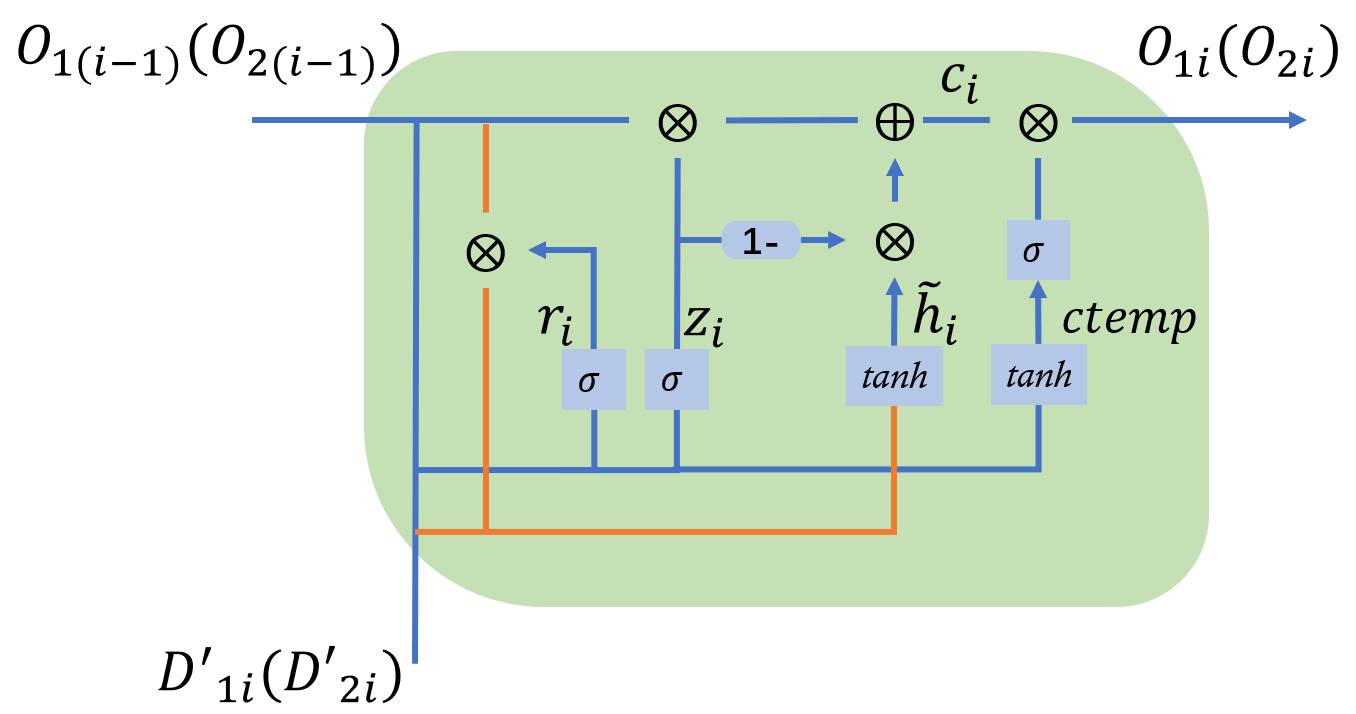}
  \caption{The multi-gated memory cell in DCMNN.}
  \label{lstmcell}
\end{figure}

Because of the poor performance and slow convergence of the original GRU in gait classification, the memory cell in DCMNN is designed to include a new path and a temporary state so as to extract more discriminative features. The memory cell is called multi-gated memory cell as shown in Fig. \ref{lstmcell}.

\begin{equation}
{ctemp} = \tanh({W_{ctemp}}\cdot[{O_{1(i-1)}},{D'_{1i}}])
\label{impv_ctemp}
\end{equation}
\vspace{-0.5cm}
\begin{equation}
{c_i} = (1-{z_i})\otimes{{\tilde h}_i}+{z_i} \otimes {O_{1(i-1)}}
\label{impv_c_i}
\end{equation}
\vspace{-0.5cm}
\begin{equation}
{O_{1i}} = {c_i}\otimes \sigma({ctemp})
\label{impv_h_i}
\end{equation}
where $ctemp$ represents the temporary state in order to determine $D'_{1i}$ and $O_{1(i-1)}$ while $c_i$ denotes the final state of the original GRU. We then extract features from $c_i$. $O_{1i}$ indicates the actual activation of the proposed node at time step $i\in\{1,2,3...,T\}$. Empirical results show that the proposed cell structure has better convergence than the standard LSTM.

We witness that the new path in the multi-gated memory cell can capture the state of the previous node and the input of the current node at the same time. $ctemp$ restricts these two states to $[-1,1]$, which highlights the difference of feature changes. Then, these differences are converted to probability values by a $sigmoid$ function, that is, the more prominent the difference is, the greater the probability is, where the range falls between 0 and 1. The final output of $O_{1i}$ successfully expresses the salient feature of the state of $c_i$.

\subsubsection{The Correlation Computing Unit}

\begin{figure}[!htb]
  \centering
  \includegraphics[width=4.5cm]{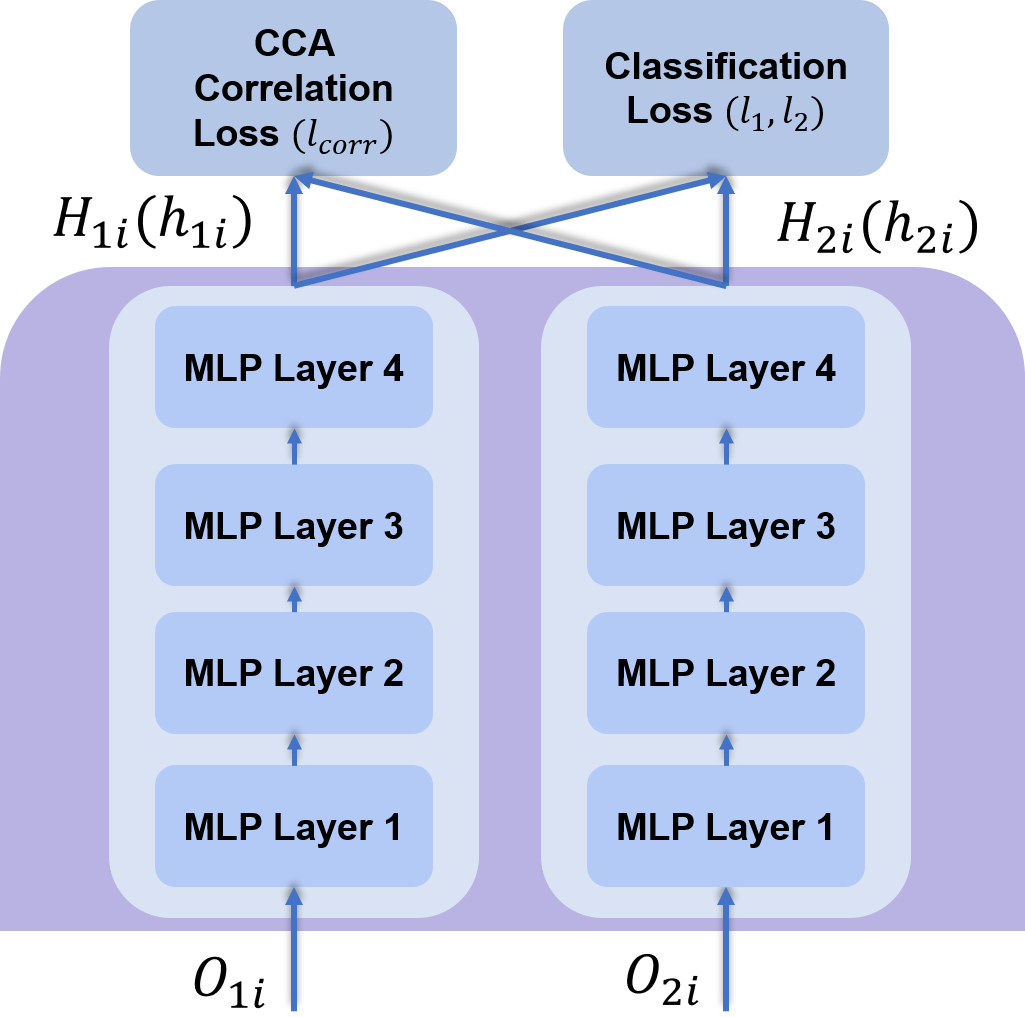}
  \caption{The correlation computing unit. We design a deep model including two four-layer MLP Networks, which can update the weights and other parameters by minimizing the losses ($l_{total},l_{1},l_{2},l_{corr}$). For the different settings of the classification and correlation analysis in MLP, the final output is set to $C$-dimension (class number) for classification ($h_{2i}, h_{1i}$) and 10-dimension for correlation calculation ($H_{2i}, H_{1i}$). Using the output of the two MLP networks, we compute the correlation of the two outputs, while the classification loss is also calculated.}
   \label{dcca}
\vspace{-0.2cm}
\end{figure}


This section introduces the basic structure of the correlation computation unit including the transmission mode of dual modal data and the operating mechanism. In this module, deep canonical correlation analysis (DCCA) \cite{Andrew2013Deep}, a correlation analysis method focusing on data fusion and multi-modal recognition is involved. It embeds deep neural networks to learn the feature representation from different modalities.


As shown in Fig. \ref{dcca}, there are three modules in our designed correlation computing unit, i.e. multi-layer perceptron (MLP) network, correlation computation and a $softmax$ mapping function. From the output of a node of DCMNN, we observe that data source one $O_{1i}$ and data source two $O_{2i}$ are simultaneously sent to the four-layer MLP unit. During the period of training, the correlation coefficient ($corr$) is calculated by CCA and its inverse ($-corr$) is minimized as the final result of the loss function. In other words, the two features extracted by MLP are relevant. Moreover, the class labels of $O_{1i}$ and $O_{2i}$ are also fed to this network to identify their corresponding categories by minimizing the classification loss. We use Adam optimizer to optimize the product of the classification and correlation losses. Eq. (\ref{losscca}) illustrates our calculation method for the losses. In conclusion, we calculate two kinds of losses to obtain feature vectors from the correlation computing cell with the best inter-data fusion and intra-data separability. Finally, the output of units $H_{1}$ and $H_{2}$ is adopted by the multi-switch discriminator.

\begin{equation}
{l_{total}} = {l_{1}}+{l_{2}}+l_{corr}
\label{losscca}
\end{equation}

\subsection{The Multi-Switch Discriminator}
As mentioned above, gaits are represented by a series of feature vectors through SFE and CorrMNN, each of whose elements is extracted as a spatio-temporal feature. The sequence is coded into the selective switch, i.e. hidden Markov model (HMM). In practice, each node processes certain features during a time period, and establishes a model to describe and code these features. The discriminator learns the internal rules of the time series data, convert the data into continuous hidden state nodes and fit the overall input data with internal parameters. Our method stems from the one used for Parkinson's disease identification \cite{khorasani2014hmm}. Compared with LSTM, our approach is a fully automated approach, which has considerable advantages in parameter estimation and computation speeds \cite{Zheheng2019Context}.

To describe each state of a dynamic system, it is not sufficient to use only one HMM switch to represent different diseases or disease severity. In our work, we use $C$ switches (referring to $C$ classes for classification) to fit the fused features, where the temporal sequence is considered as hidden state nodes. Upon the arrival of a gait sample at this module, it analyzes the scores of the four HMM switches to determine which HMM they belong to and subsequently determine the output category (i.e. PD, HD, ALS or CO) as the final result of the system. 

\begin{table*}
\centering
\caption{Performance of the proposed approach in classifying two groups (CO and NDDs), compared with the other methods. "-" means not applicable.}
\scalebox{1}{
\begin{tabular}{lcccc}
  \hline
  \hline
 Single-modal Methods&ALS vs. CO&PD vs. CO&HD vs. CO&NDDs vs. CO\\
  \hline
  RBF+DL \cite{Zeng2015Classification}&89.66\%&87.10\%&83.33\%&-\\
  QBC \cite{Banaie2011Introduction}&100\%&80.00\%&71.43\%&-\\
  Meta-classifiers \cite{S2014Gait}&96.13\%&90.36\%&88.67\%&-\\
  HMM \cite{khorasani2014hmm}&-&90.32\%&-&-\\
  C-FuzzyEn+SVM \cite{Yi2016Symmetry}&-&96.77\%&-&-\\
  PE+SVM \cite{Xia2016A}&92.86\%&-&-&-\\
 \hline
 Multi-modal Methods&ALS vs. CO&PD vs. CO&HD vs. CO&NDDs vs. CO\\
  \hline
  DCLSTM \cite{Zhao2018Dual}&97.43\%&97.33\%&94.96\%&96.42\%\\
  The proposed method&\textbf{\textit{100\%}}&\textbf{\textit{99.86\%}}&\textbf{\textit{99.74\%}}&\textbf{\textit{99.47\%}}\\
\hline
\hline
\end{tabular}}
  \label{tb_2class}
\end{table*}

\begin{table}[!htb]
\centering 
\caption{Performance of classifying NDDs (ALS,HD,PD).}
\scalebox{1}{
\begin{tabular}{l|ll}
  \hline
  \hline
  Single-modal Methods&\multicolumn{2}{c}{Multi-modal Methods}\\
  \hline
 QBC \cite{Banaie2011Introduction}&DCLSTM \cite{Zhao2018Dual}&The proposed method\\
  \hline
  86.96\%&95.67\%&\textbf{\textit{98.88\%}}\\
\hline
\hline
\end{tabular}}
  \label{tb_3class}
\end{table}

\section{Experiments}

This section presents our experimental setup and the results of the proposed method, compared against several state of the art methods on three challenging gait datasets.

\subsection{Experimental Settings}
\subsubsection{Dataset Specifications}
In this section, we give a brief description of the three datasets used in our experiment. The multimodal data used in our experiment is shown in Fig. \ref{data}.

\textbf{NDDs dataset}: This is a dual-modal dataset that contains two different types of gait data, i.e. temporal and force-sensing data of 48 patients with NDDs (ALS, HD and PD) and 16 healthy controls (CO)  \cite{physionet-ndd}. The dataset was collected simultaneously using different sensors in a foot-switch system, which provided accurate estimates of the starting and the finishing stages of successive steps using commercially available transducers (this data collection can be easily reproduced in a laboratory environment). Moreover, the hardware includes two 1.5 $in^2$ force sensing resistors and a 390$\Omega$ measuring resistor, which can acquire the stride time intervals by measuring the variation tendency of the force changes in gait \cite{Hausdorff1995Footswitch,Hausdorff1997Altered,Hausdorff2000Dynamic}.


\textbf{PDgait dataset}:
This is a reorganized gait dataset of Parkinson's disease (PD), including two sub-dataset, one of which is the vertical ground reaction force sequence (VGRF) collected by the force sensor, and the other is the acceleration data collected by the acceleration sensor. We combine these two sets of data to form the gaits of PD patients with different severity degrees, referring to the Hoehn \& Yahr standard.

The VGRF database contains the measures of gaits from 93 patients with idiopathic PD as they walked at their usual and self-selected paces for approximately 2 minutes on level ground. Underneath each foot were 8 sensors that measure force (in Newtons) as a function of time. The output of each of these 16 sensors has been digitized and recorded at 100 samples per second \cite{Goldberger}. This database also includes demographic information, measures of disease severity and other related measures.

The acceleration database contains the annotated measures of 3 acceleration sensors attached to hips and legs of PD patients that experience gait disorders during walking tasks \cite{5325884}. The data was collected in the lab, and the subjects performed three tasks: straight line walking, walking with numerous turns, and finally more realistic activities of daily living.

\textbf{SDUgait dataset}: The dataset includes 52 subjects, each of who has 20 sequences with at least 6 fixed walking directions and 2 arbitrary directions, and totally 1040 sequences. Two Kinects were used for simultaneously capturing the 3D positions of 21 joints and corresponding binarized silhouette images after people subtracted the foreground from the background using depth images collected at each frame. We aim to recognize a normal gait by using 3D skeleton nodes and 2D gait silhouettes \cite{7532940}.

\subsubsection{Set-Up}
The experiments are conducted with Tensorflow \cite{Tensorflow} and Python \cite{Python} libraries. We will introduce the parameter settings for the three modules: SFE, CorrMNN and multi-switch determinator on the three datasets.

In the training process, the computation complexity of the three modules is $O(K(2D+1))$ (SFE), $O(4(NM+N^2+N))$ (CorrMNN) and $O(2N^2T)$ (multi-switch discriminator), where $N$ is the number of the hidden states and $T$ is the length of the observation sequence respectively, determined by the training parameters of each module. In addition, for testing, the mean execution time for achieving the best experimental results of the three modules is 0.378min (SFE), 0.718s (CorrMNN), 0.512s (multi-switch discriminator), using the system of GTX1050Ti GPU, i5-7300HQ CPU and 8G RAM.

For the NDDs dataset, in the SFE module, the number of the Gaussian distributions of the GMM model $K$ is set to 15 for the temporal data and 20 for the force-sensing data due to its high dimensionality. In the CorrMNN module, the normalized data enters the DCMNN in the form of $timestep \times feature~number$ for each training sample. $timestep$ and $feature~number$ of the time series channel are 10 (10 frames containing the information of 10 seconds and 10 nodes in DCMNN) and 12 (12-dimension feature in one frame) respectively, while the numbers of the force-sensing channel are set to 10 (containing the information of 10 seconds) and 60$\times$150 (60-dimension feature in one frame and 150 frames in one training sample) respectively. In addition, with a learning rate of 0.01 and a batch size of 256, the DCMNN model is trained using a hidden output dimensionality of 256 in each DCMNN node. In the multi-switch discriminator, it includes four HMM switches (ALS, HD, PD and CO). The settings include the number of the hidden states $N=10$ (to be associated with the DCMNN internal nodes containing the temporal and fused feature of 10 seconds), and the number of the training iterations $I=200$.

\begin{table}[!htb]
\centering 
\caption{Performance of classifying four groups (ALS,PD,HD,CO) compared with other models.}
\begin{tabular}{lc|lc}
  \hline
  \hline
  Single-modal Methods&Accuracy&Multi-modal Methods&Accuracy\\
  \hline
  CapsNet&88.86\%&MELD-LSTMs \cite{Poria2018MELD}&88.85\%\\
  HMM&83.94\%&RC-LSTMs\cite{Filippo2016Human}&90.76\%\\
  Original LSTM&93.42\%&DCLSTM \cite{Zhao2018Dual}&95.84\%\\
  CNN&85.58\%&DCCA&68.45\%\\
  GRU&93.61\%&KCCA&65.78\%\\
  BiLSTM&95.59\%&The proposed method&\textbf{\textit{99.31\%}}\\
\hline
\hline
\end{tabular}
  \label{advanced_model}
\end{table}

\begin{figure}
     \centering
     \begin{subfigure}[b]{0.24\textwidth}
         \centering
         \includegraphics[width=4.5cm]{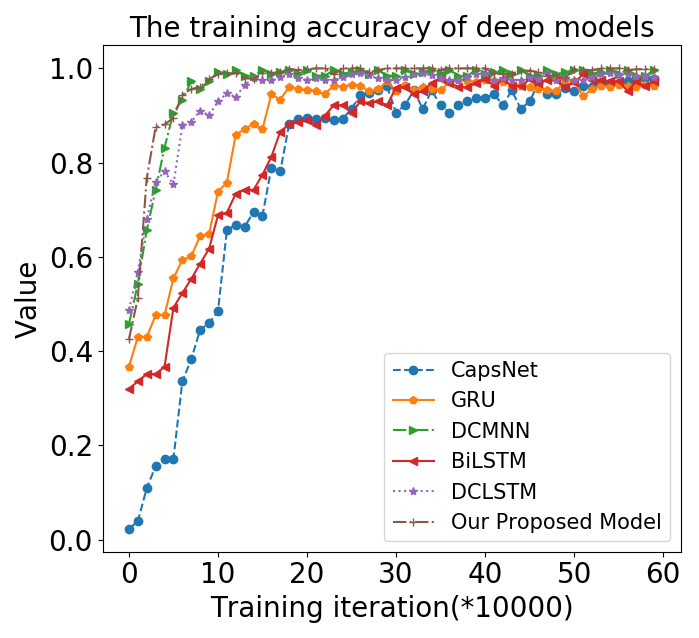}
         \caption{The training accuracy of deep models.}
         \label{acc_loss:acc}
     \end{subfigure}
     \begin{subfigure}[b]{0.24\textwidth}
         \centering
         \includegraphics[width=4.5cm]{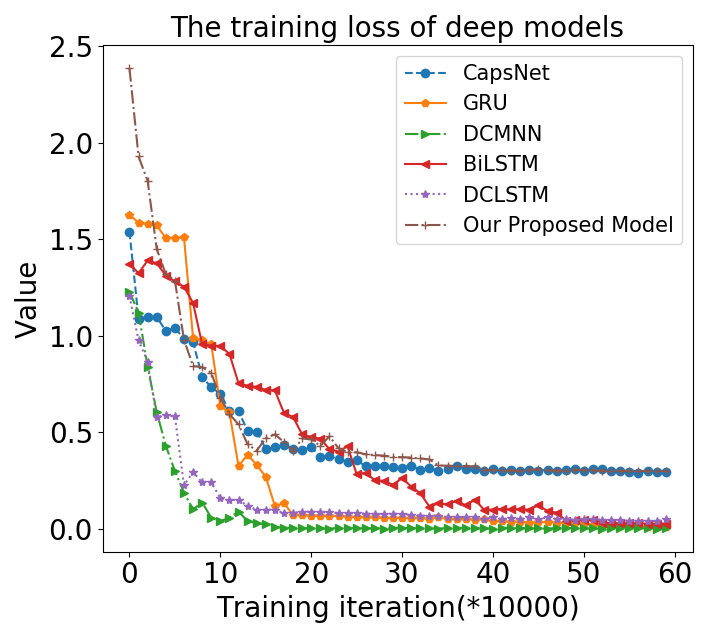}
         \caption{The training loss of deep models.}
         \label{acc_loss:loss}
     \end{subfigure}
\caption{The training accuracy and loss of different deep models on the NDDs dataset. In sub-figure \subref{acc_loss:acc}, the first three models with the best convergence rate are DCMNN, DCLSTM and our proposed model. In addition, our proposed model is the best one with the highest classification accuracy. In sub-figure \subref{acc_loss:loss}, DCMNN and DCLSTM converge faster and tend to be stable. Our proposed model uses a joint loss function with a large initial value, but finally the convergence value is less than 0.5.}
\label{acc_loss}
\end{figure}

\begin{figure*}
     \centering
     \begin{subfigure}[b]{0.23\textwidth}
         \centering
         \includegraphics[width=4.6cm,height=4.4cm]{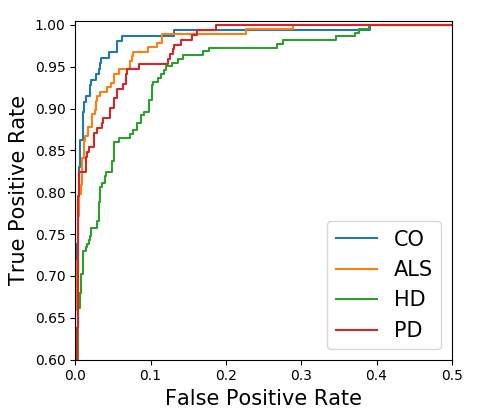}
         \caption{ROC curve for Fisher Vector.}
         \label{roc:fv}
     \end{subfigure}
     \hfill
     \begin{subfigure}[b]{0.23\textwidth}
         \centering
         \includegraphics[width=4.6cm,height=4.7cm]{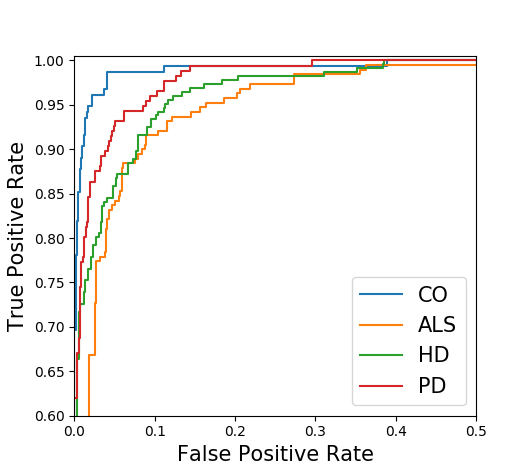}
         \caption{ROC curve for LDA.}
         \label{roc:lda}
     \end{subfigure}
     \hfill
     \begin{subfigure}[b]{0.23\textwidth}
         \centering
         \includegraphics[width=4.6cm,height=4.3cm]{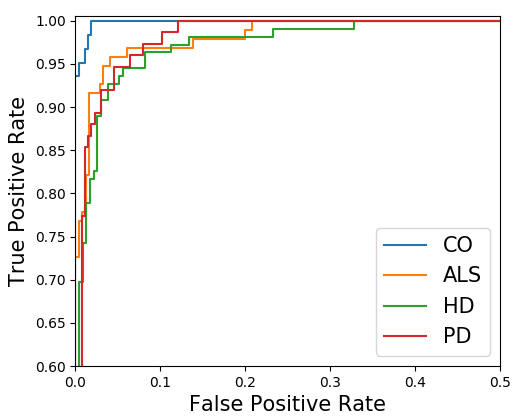}
         \caption{ROC curve for SFE.}
         \label{roc:sfe}
     \end{subfigure}
     \hfill
     \begin{subfigure}[b]{0.26\textwidth}
         \centering
         \includegraphics[width=4.8cm,height=4.3cm]{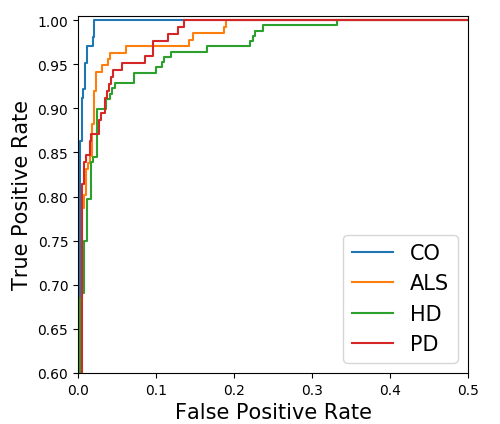}
         \caption{ROC curve for DCMNN.}
         \label{roc:dclstm}
     \end{subfigure}
     \vfill
     \begin{subfigure}[b]{0.23\textwidth}
         \centering
         \includegraphics[width=4.6cm,height=4.4cm]{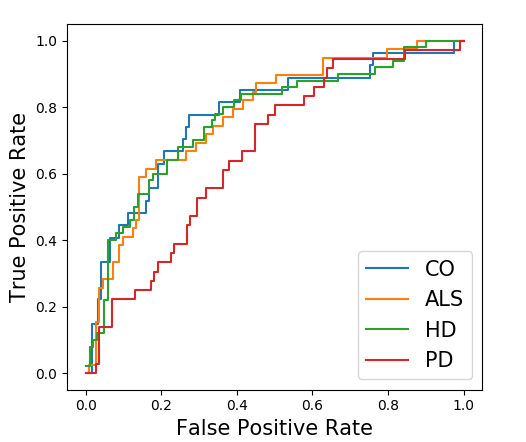}
         \caption{ROC curve for DCCA.}
         \label{roc:dcca}
     \end{subfigure}
     \vspace{0.5cm}
     \hfill
     \begin{subfigure}[b]{0.23\textwidth}
         \centering
         \includegraphics[width=4.6cm,height=4.3cm]{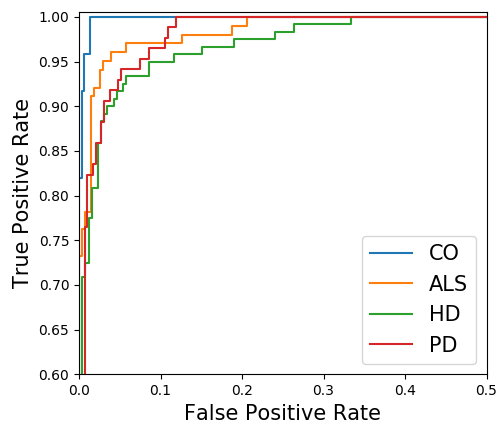}
         \caption{ROC curve for CorrMNN}
         \label{roc:CorrMNN}
     \end{subfigure}
     \hfill
     \begin{subfigure}[b]{0.23\textwidth}
         \centering
         \includegraphics[width=4.6cm,height=4.4cm]{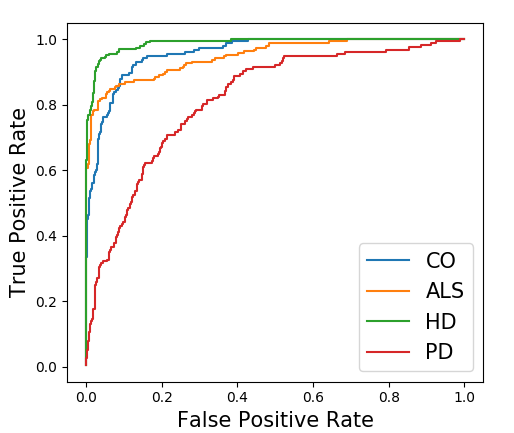}
         \caption{ROC curve for HMM.}
         \label{roc:hmm}
     \end{subfigure}
     \hfill
     \begin{subfigure}[b]{0.26\textwidth}
         \centering
         \includegraphics[width=4.8cm,height=4.3cm]{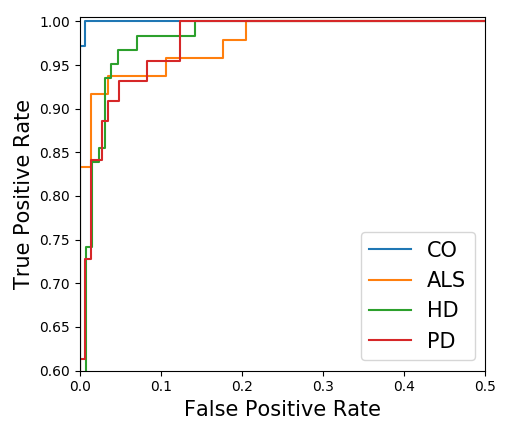}
         \caption{ROC curve for the whole model.}
         \label{roc:fusion}
     \end{subfigure}
\caption{ROC curves show that the true positive rate (TPR) against the false positive rate (FPR) for the features extracted by different components: \subref{roc:fv} Fisher Vector features in SFE; \subref{roc:lda} LDA; \subref{roc:sfe} SFE; \subref{roc:dclstm} DCLSTM; \subref{roc:dcca} DCCA; \subref{roc:CorrMNN} CorrMNN; \subref{roc:hmm} HMM; \subref{roc:fusion} all the combined features.}
\label{roc}
\end{figure*}

For the PDgait dataset, in the SFE module, $K$ of the GMM is defined as 15 for the VGRF data and 10 for the acceleration data. In the CorrMNN module, $timestep$ and $input~dimension$ of the VGRF channel are 100 and 19 while those of the acceleration channel are set to 100 and 9. This experiment has 3 switches for 3 different severity levels (Severity 2, Severity 2.5, Severity 3). We define the number of the hidden states as $N=100$ with 500 training iterations.

For the SDUgait dataset, experimentally, the value of $K$ in the SFE module is set to 10 for the 3D skeleton data and 20 for the image data. In the CorrMNN module, $timestep$ and $input~dimension$ of the 3D skeleton channel are 10 and 63 respectively while those of the image channel are set to 10 and 2500, respectively. This experiment has 52 switches (Subject 1, Subject 2, Subject 3, ..., Subject 52). We define the number of the hidden states as $N=10$ with 500 training iterations.

\begin{table*}[!htb]
\centering 
\caption{Comparison of different components on the NDDs dataset.}
\scalebox{1}{
\begin{tabular}{lccccccccc}
  \hline
  \hline
  Disease Class&Fisher Vector&LDA&SFE&DCMNN&DCCA&CorrMNN&HMM&Combined&Mean\\
  \hline
  CO&93.59\%&91.39\%&95.83\%&96.23\%&70.54\%&100\%&87.17\%&100\%&91.84\%\\
  ALS&93.21\%&87.24\%&93.75\%&95.56\%&69.33\%&96.55\%&83.72\%&100\%&89.92\%\\
  HD&89.61\%&89.58\%&95.83\%&95.75\%&67.44\%&98.34\%&82.76\%&100\%&89.91\%\\
  PD&90.03\%&90.32\%&93.75\%&95.82\%&66.49\%&98.46\%&82.11\%&97.22\%&89.28\%\\
  All&\textbf{\textit{91.61\%}}&\textbf{\textit{89.63\%}}&\textbf{\textit{94.79\%}}&\textbf{\textit{95.84\%}}&\textbf{\textit{68.45\%}}&\textbf{\textit{98.34\%}}&\textbf{\textit{83.94\%}}&\textbf{\textit{99.31\%}}\\
\hline
\hline
\end{tabular}}
  \label{components_ndds}
\end{table*}

\begin{figure}
     \centering
     \begin{subfigure}[b]{0.24\textwidth}
         \centering
         \includegraphics[width=4.5cm]{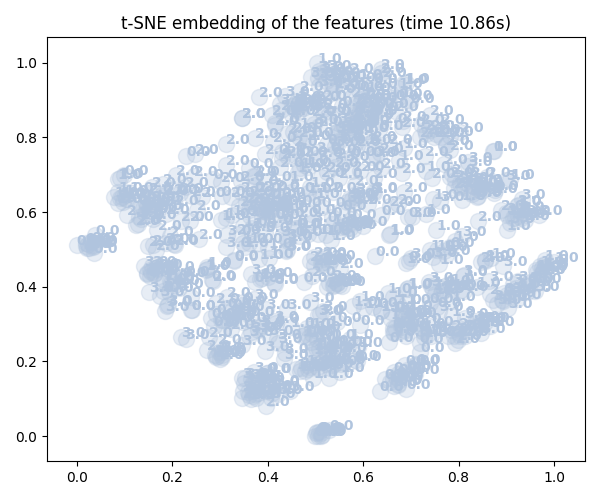}
         \caption{Raw data}
         \label{ndds_fea:org}
     \end{subfigure}
     \begin{subfigure}[b]{0.24\textwidth}
         \centering
         \includegraphics[width=4.5cm]{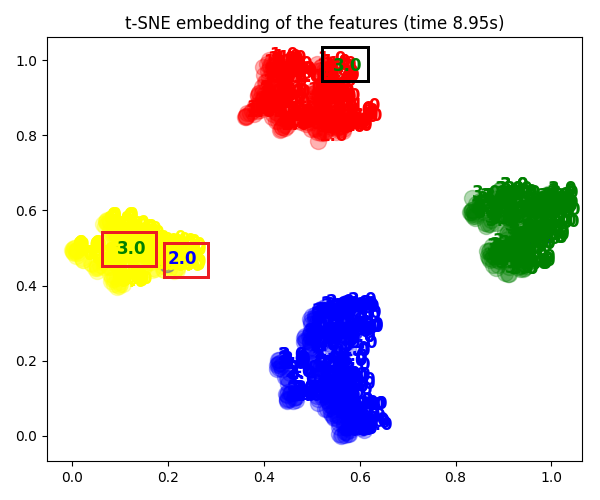}
         \caption{The extracted feature}
         \label{ndds_fea:fea}
     \end{subfigure}
\caption{Comparison of t-SNE feature dimensionality reduction on the NDDs dataset. Sub-figures \subref{ndds_fea:org} and \subref{ndds_fea:fea} represent the comparison between the original data and the feature extracted from our model after dimensionality reduction has been applied on the NDDs dataset. We have also highlighted some examples of misclassification.}
\label{ndds_fea}
\end{figure}

\begin{table}[!htb]
\centering 
\caption{Comparison of several state of the art methods in distinguishing three severity levels (Severity 2, Severity 2.5, Severity 3) of Parkinson's disease on the PDgait dataset.}
\begin{tabular}{lc|lc}
  \hline
  \hline
  Single-modal Methods&Accuracy&Multi-modal Methods&Accuracy\\
  \hline
  CapsNet&95.01\%&DCCA&94.30\%\\
  HMM&92.20\%&KCCA&92.48\%\\
  Q-BTDNN\cite{jane2016q}&93.10\%&DCLSTM\cite{Zhao2018Dual}&96.71\%\\
  Original LSTM&91.13\%&The proposed method&\textbf{\textit{98.93\%}}\\
  CNN&82.86\%&&\\
  LSTM+CNN\cite{ZHAO20181}&97.48\%&&\\
  GRU&92.53\%&&\\
  BiLSTM&91.58\%&&\\
\hline
\hline
\end{tabular}
  \label{advanced_model_park}
\end{table}

\begin{table*}[!htb]
\centering 
\caption{Comparison of different components on the PDgait dataset.}
\scalebox{1}{
\begin{tabular}{lccccccccc}
  \hline
  \hline
  Severity Level&Fisher Vector&LDA&SFE&DCMNN&DCCA&CorrMNN&HMM&Combined&Mean\\
  \hline
  Severity 2&91.51\%&91.84\%&93.49\%&97.33\%&95.13\%&97.93\%&92.82\%&99.15\%&94.92\%\\
  Severity 2.5&84.24\%&68.47\%&92.49\%&90.89\%&93.26\%&95.28\%&89.53\%&98.46\%&89.08\%\\
  Severity 3&90.68\%&86.15\%&94.27\%&97.02\%&93.98\%&95.20\%&93.24\%&98.99\%&93.69\%\\
  All&\textbf{\textit{89.53\%}}&\textbf{\textit{84.45\%}}&\textbf{\textit{93.52\%}}&\textbf{\textit{95.08\%}}&\textbf{\textit{94.30\%}}&\textbf{\textit{96.13\%}}&\textbf{\textit{92.20\%}}&\textbf{\textit{98.93\%}}\\
\hline
\hline
\end{tabular}}
  \label{components_park}
\end{table*}

\subsection{The Results on NDDs dataset}

There are 63 participants in the dataset after we have removed one HD patient due to the data confusion and noise, including 16 healthy controls (CO), 13 ALS patients, 15 PD patients and 19 HD patients. We divide them into different groups according to the experimental settings.

\subsubsection{Classification of Patients and Healthy People}

The first experiment consists of four comparison groups: ALS vs. CO, PD vs. CO, HD vs. CO, and NDDs vs. CO, which show the difference between healthy people and NDDs patients. Since the data is divided into two categories, only two HMM switches (one of ALS, HD, PD switches and CO) are used to model the engaged data. The performance of the proposed model is shown in Table \ref{tb_2class}. In this work, we use 80\% of the data for training and 20\% for testing. The leave-one-out cross-validation method in sklearn library can also be applied to verifying the classification accuracy of the proposed model. For ALS detection, we can find that all the comparison methods are more than 90\%, except for the deterministic learning method RBF+DL \cite{Zeng2015Classification}, due to the weakness of using only single modal data. The average performance of deep models is superior than that of the classifiers like PE+SVM \cite{Xia2016A} and C-FuzzyEn+SVM \cite{Yi2016Symmetry}. QBC \cite{Banaie2011Introduction} achieves 100\% of the result which benefits for the generated novel feature of the gait data. For the detection of PD patients, the results of multi-modal method are much better than that of single-modal methods. The features extracted by C-FuzzyEn+SVM \cite{Yi2016Symmetry} are more representative outperforming other single-modal methods. For HD detection, the compared methods are very poor due to the similarity between HD and other diseases, but the multi-modal methods still show their advantages. The dual channel LSTM (DCLSTM) \cite{Zhao2018Dual} achieves better result among these studies by fusing the two types of data and capture the temporal characteristics. Compared with the four different groups, we observe that the proposed method can distinguish the ALS and CO groups effectively while having worse performance on NDDs vs CO group.


\subsubsection{Classification of Different Diseases}

\begin{figure}
     \centering
     \begin{subfigure}[b]{0.24\textwidth}
         \centering
         \includegraphics[width=4.6cm]{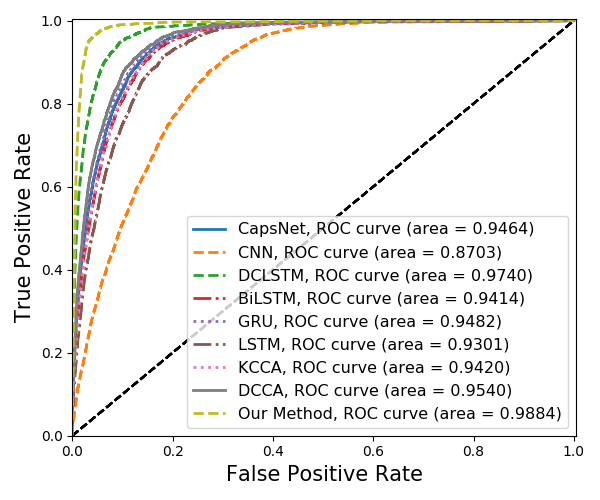}
         \caption{Advanced models.}
         \label{roc_compare:model_park}
     \end{subfigure}
     \begin{subfigure}[b]{0.24\textwidth}
         \centering
         \includegraphics[width=4.6cm]{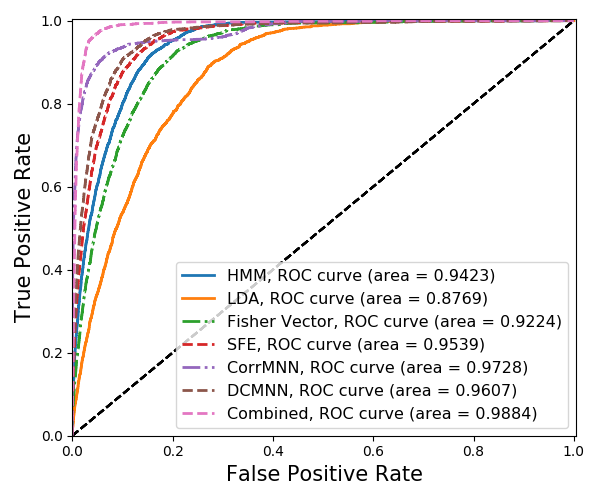}
         \caption{Constituting components.}
         \label{roc_compare:comp_park}
     \end{subfigure}
\caption{Performance of the advanced models and constituting components of the proposed model on the PDgait dataset. Sub-figures \subref{roc_compare:model_park} and \subref{roc_compare:comp_park} represent the performance of the advanced models and constituting components of the proposed model on the PDgait dataset.}
\label{roc_compare_fog}
\end{figure}

The second experiment is designed to separate 47 patients with NDDs (ALS, HD and PD) using their gait patterns for identifying the subtle distinctions of symptoms. We adopt three HMM switches to classify the fusion correlation features from CorrMNN. Similar to the first experiment, we have 20\% of the data for testing. Table \ref{tb_3class} shows the classification accuracy of the three studies. Compared with the first experiment, the number of the categories is increased, leading to a slight decrease in recognition accuracy. Banaie \textit{et al.} \cite{Banaie2011Introduction} used QBC to obtain the best result (86.96\%) which did not consider the temporal changes of gaits and these traditional classifiers are not consistent. The dual channel LSTM (DCLSTM) \cite{Zhao2018Dual} and our proposed method are capable of extracting more representative features so the classification results seem better.

\subsubsection{Classification of All the States of Subjects}

In this subsection, we will validate the performance of our proposed framework in identifying all the states. Given four classes (ALS,PD,HD,CO), four HMM switches are used to fit the output data of the CorrMNN and the other settings of the parameters in the model remain unchanged. Table \ref{advanced_model} shows the performance of several state of the art methods. Because the traditional classifiers do not take into account complex temporal and spacial information, we do not compare these methods here.

A HMM uses the temporal information in its hidden state when handling continuous streaming signals, and therefore it is not suitable to distinguish uncorrelated signals with the accuracy of less than 90\%. The results of CNN and CapsNet are similar, because CapsNet is an upgraded CNN network, which mainly analyzes spatial information of the gait. The proposed hybrid model significantly outperforms the other state-of-the-art single-modality methods. In general, the temporal feature extraction model outperforms the spatial feature extraction model, and the classification result of BiLSTM in LSTM related models reaches 95.59\%.

In addition, RC-LSTMs \cite{Filippo2016Human} and MELD-LSTMs \cite{Poria2018MELD} include different combinations of LSTMs for feature extraction, using the previous gait patterns to predict the current pattern. DCCA and KCCA can not be independently used due to the limitation of internal simple structure. However, these methods do not analyze the signals in time and frequency domains at the same time. In addition, when they extract features from multi-modal data, the correlation between different data, the dependence on a large number of preprocessing steps and specific settings are ignored.


\begin{figure}
     \centering
     \begin{subfigure}[b]{0.24\textwidth}
         \centering
         \includegraphics[width=4.5cm]{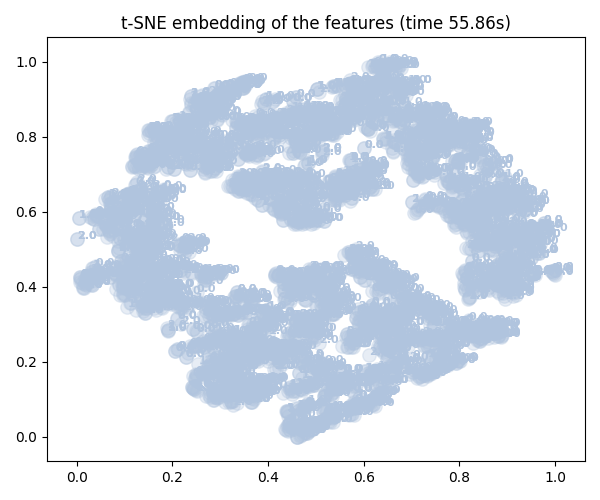}
         \caption{Raw data.}
         \label{feature_compare:raw_park}
     \end{subfigure}
     \begin{subfigure}[b]{0.24\textwidth}
         \centering
         \includegraphics[width=4.5cm]{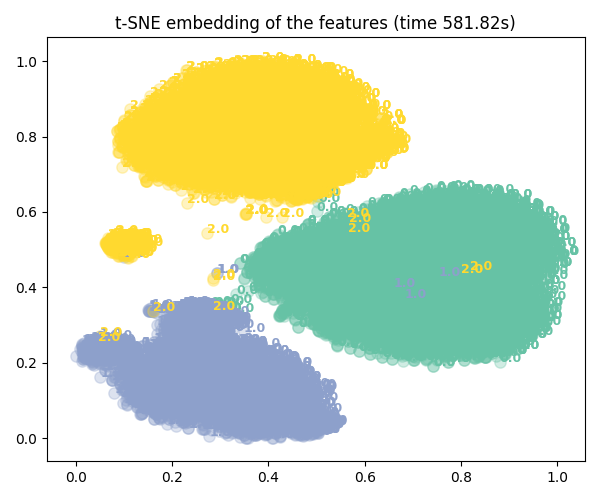}
         \caption{The extracted feature.}
         \label{feature_compare:feature_park}
     \end{subfigure}
\caption{Comparison of t-SNE feature dimensionality reduction on the PDgait dataset. Sub-figures \subref{feature_compare:raw_park} and \subref{feature_compare:feature_park} represent the comparison between the original data and the feature extracted from our model after dimensionality reduction on the PDgait dataset.}
\label{feature_compare_fog}
\end{figure}

\subsubsection{Performance of the Entire Model}
In this section, taking the third experiment as an example (for four groups classification), we focus on investigating the effect of the method described in Section III. Fig. \ref{acc_loss} demonstrates the accuracy and the cost of the training process per iteration by different deep models.

In Fig. \ref{acc_loss:acc}, the accuracy of the CapsNet classification starts at a lower rate than the other models, and the convergence is slower because it takes into account both the logical and structural information of data. Other deep models aim to learn the temporal information of the input sequences, which start from 0.35. The single modality model GRU, bidirectional LSTM (BiLSTM) and CapsNet do not perform as well as the three multi-modal methods i.e. DCMNN, DCLSTM and our proposed model.

Fig. \ref{acc_loss:loss} illustrates the comparison of the training losses of these deep models in the recognition process. Our proposed model and CapsNet result in a figure larger than those of the other methods because our proposed model used a joint loss function, while CapsNet has slow convergence with different loss functions. Compared to the other four methods, BiLSTM is of the slowest convergence after the $30^{th}$ iteration because of the complex internal structure of its bidirectional computing nodes, and the DCMNN and DCLSTM are of the lowest cost.

Additionally, we compare the recognition performance of each part of the hybrid model using receiver operating characteristic curve (ROC). In Fig. \ref{roc}, we can see that the best performance and the fastest convergence is (h). When FPR = 0.2, the TPR of the four categories reaches nearly 100\%. In contrast, (e) is the worst model which does not use temporal features of the gait data. (c) and (g) perform poorly, because the two models do not take advantage of temporal features, so the temporal models are more suitable for the gait recognition task, such as (d) and (f). Although (a) and (b) do not take into account temporal features, the fusion of Fisher Vector and LDA for the powerful ability of spatial feature extraction and classification leads to good results in (c). Moreover, compared with the four classes, CO is the most accurate class to be identified with the lowest error rate and the fastest convergence. Other classes have varying degrees of volatility.

Table \ref{components_ndds} shows the recognition accuracy of each component of the hybrid model. The row here represents the recognition rate of each sub-model for a certain state, and the column represents the recognition results of a sub-model for different states. Compared with the performance of sub-models, the recognition rate of the fusion model is the highest. It is worth noting that the best testing result is that the hybrid model can completely recognize the gait of CO, HD and ALS, indicating that the differences among the three gaits are evident, which may be affected by the inconsistency of subjects' biological characteristics. Furthermore, by calculating the average accuracy of all submodels in each class, the results are CO (91.84\%), and the worst is PD (89.28\%) while ALS (89.92\%) and HD (89.91\%) achieve similar results. It can be seen that the model needs to be further improved for PD recognition. 
\begin{figure}
     \centering
    \begin{subfigure}[b]{0.24\textwidth}
         \centering
         \includegraphics[width=4.6cm]{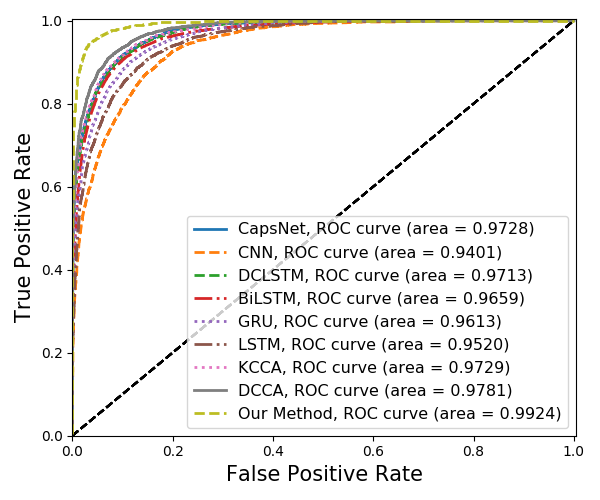}
         \caption{Advanced models.}
         \label{roc_compare:model_sdu}
     \end{subfigure}
     \begin{subfigure}[b]{0.24\textwidth}
         \centering
         \includegraphics[width=4.6cm]{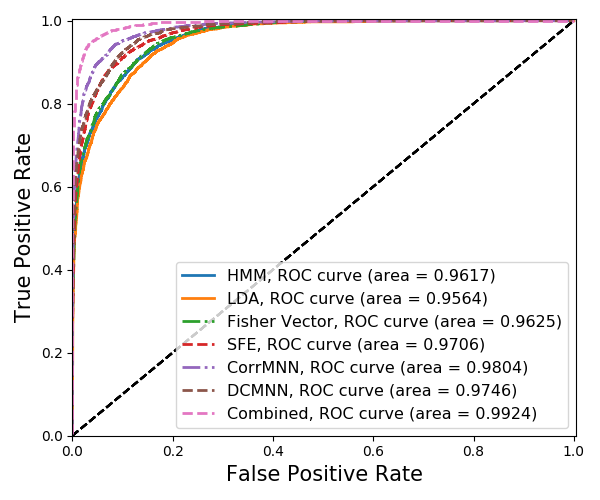}
         \caption{Constituting components.}
         \label{roc_compare:comp_sdu}
     \end{subfigure}
\caption{Performance of several advanced models and the constituting components of the proposed model on the SDUgait dataset. Sub-figures \subref{roc_compare:model_sdu} and \subref{roc_compare:comp_sdu} represent the performance of the advanced models and the constituting components of the proposed model on the SDUgait dataset.}
\label{roc_compare_sdu}
\end{figure}
Compared with all the components in the proposed model, the purpose of Fisher Vector is to extract the spatial information of gait data. LDA reduces the dimension of Fisher Vector features and classifies them to obtain separable spatial gait features. The purpose of DCMNN is to analyze the temporal information of gait. DCCA is used to connect two channels to calculate their correlation, and a fusion model CorrMNN is obtained to generate fused temporal features. HMM is used as a decision-making layer to generate accurate predicted labels based on spatio-temporal features.

The effect of LDA and Fisher Vector is similar, while SFE has a slight increase to verify the validity of the model fusion. So is CorrMNN component, which is more efficient than the fusion of DCCA and DCMNN. HMM is placed in the last layer, using the fusion features of CorrMNN and SFE, and the result is much better than using HMM independently.

Additionally, we can see the CorrMNN module achieves the best result 98.34\%, compared with the other constituting components in the proposed model, which is only second to the whole model for NDDs classification. Although the results of SFE and HMM cannot catch up with those of temporal models such as LSTM, their combination is higher than individual accuracy, and also exceeds the results of CorrMNN. Therefore, the hybrid model of these three modules is selected for feature extraction and classification.


Moreover, with the help of the t-SNE visualization method, the classification ability of our proposed method is shown in Fig. \ref{ndds_fea}. We can see that the extracted features have very few errors. The three error cases are caused by the high similarity between the testing signal sequence and the other signals. The gait samples with labels 2 and 3 (i.e., HD and PD) are misclassified into the classes with labels 0 and 1 (i.e., CO and ALS). By comparing the gait features of these classes, it is found that the high similarity of the signals leads to wrong classification results. The same situation also happens on the other two datasets (i.e., PDgait and SDUgait), which will be omitted here. 

\begin{figure}
     \centering
    \begin{subfigure}[b]{0.24\textwidth}
         \centering
         \includegraphics[width=4.5cm]{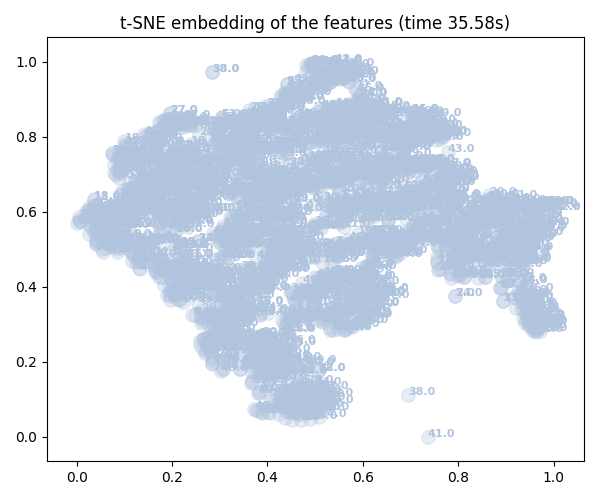}
         \caption{Raw data.}
         \label{feature_compare:raw_sdu}
     \end{subfigure}
     \begin{subfigure}[b]{0.24\textwidth}
         \centering
         \includegraphics[width=4.5cm]{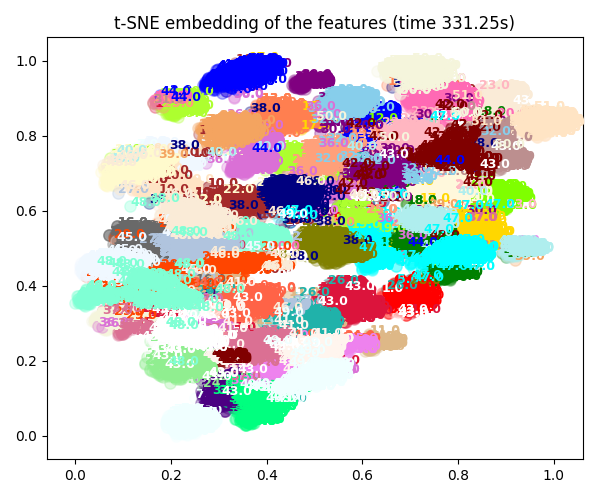}
         \caption{The extracted feature.}
         \label{feature_compare:feature_sdu}
     \end{subfigure}
\caption{Comparison of the t-SNE feature dimensionality reduction on the SDUgait dataset. Sub-figures \subref{feature_compare:raw_sdu} and \subref{feature_compare:feature_sdu} represent the comparison between the original data and the features extracted from our proposed model after dimensionality reduction on the SDUgait dataset.}
\label{feature_compare_sdu}
\end{figure}
\begin{table*}[!htb]
\centering 
\caption{Comparison of different state of the art methods and the constituting components of our proposed model in distinguishing the identities of 52 subjects.}
\begin{tabular}{lclc|lc}
  \hline
  \hline
  Single-modal Methods&Accuracy&Multi-modal Methods&Accuracy&Constituting Components&Accuracy\\
  \hline
  CapsNet&96.79\%&KCCA&94.55\%&Fisher Vector&92.81\%\\
  3D-STCNN \cite{HUYNHTHE2020}&91.04\%&DCCA&96.17\%&SFE&95.38\%\\
  Original LSTM&94.80\%&DCLSTM \cite{Zhao2018Dual}&94.64\%&DCMNN&97.26\%\\
  GRU&95.30\%&The proposed method&\textbf{\textit{98.79\%}}&DCCA&96.17\%\\
  CNN&92.85\%&&&CorrMNN&97.38\%\\
  Hand-crafted Method \cite{7532940}&77.25\%&&&HMM&95.26\%\\
  BiLSTM&96.38\%&&&LDA&90.60\%\\
\hline
\hline
\end{tabular}
  \label{advanced_model_sdu}
\end{table*}

\subsection{The Results on PDgait dataset}
\subsubsection{Classification of Different Parkinson's Severity}
This study includes more than 10 thousand training samples from the subjects with PD severity levels of 2, 2.5 and 3. The dataset is re-shaped based on the ratio of 1:1, ignoring other physiological differences between subjects and only focusing on the movement changes.

The performance of several state of the art methods are shown in Table \ref{advanced_model_park}. First of all, we divide the comparison studies into multi-modal and single-modal methods. The multi-modal methods include DCCA, KCCA, DCLSTM and our proposed method, and the rest are single-modal methods. The results of the single-modal methods come from the best results of two groups of single-modal data training respectively. In multi-modal methods, DCLSTM outperforms DCCA and KCCA, which shows the superiority of the temporal model. Among the single-modal approaches, CNN is the worst, being inferior to the temporal model. The fusion model is better, e.g. the result of the LSTM+CNN fusion model is 97.48\%. The proposed hybrid method has achieved 98.93\% accuracy that is the highest.

ROC curve demonstrates the performance of the constituting components of the proposed model and the learning models. AUC (area under curve) value is calculated as the evaluation criterion, and higher AUC indicates that the model has a stronger classification ability, shown in  Fig. \ref{roc_compare_fog}. By comparing ROC curves, we can see that the top three are multi-modal models, i.e. our proposed model, DCLSTM and DCCA. The performance of CNN is significantly lower than that of other models.

The performance of the constituting components of the proposed model for each severity level is shown in Table \ref{components_park}. LDA does not work well for recognizing PD patients with the severity of 2.5, indicating that LDA is still lacking in the calculation of maximizing the distance between classes. The output feature classification accuracy of the SFE module is higher than that of LDA and Fisher Vector, indicating the high separability of the fusion spatial features. The result of the CorrMNN module is better than that of DCMNN without correlation analysis and the spatial analysis module SFE. The average recognition rates of all the components for PD patients with severity levels 2, 2.5 and 3 are 94.92\%, 89.08\% and 93.69\%, indicating that the model has certain advantages in identifying the patients with mild and severe PD.

\subsubsection{Performance of the Whole Model}

Furthermore, we also utilize the t-SNE visualization method to show the classification ability of our proposed method. The experimental results are shown in Fig. \ref{feature_compare_fog}. Sub-figures \subref{feature_compare:raw_park} and \subref{feature_compare:feature_park} represent the comparison between the features extracted from our model and the original data after dimensionality reduction. We can clearly see that the original data is more chaotic than the extracted features, and the data of the three classes have cross coverage, while the features extracted by the proposed model have a distance between different classes and a large degree of aggregation for the same class, showing better separability.

\subsection{The Results on SDUgait dataset}
In this experiment, there are more than 20 thousand samples from 52 healthy subjects. The sample data comes from two modalities, i.e. images and 3D skeleton joints. The purpose of the experiment is to distinguish different identities based on the bimodal features of gaits.

In this experiment, we analyze the gait differences of Parkinson's disease in order to distinguish the severity level based on the Hoehn \& Yahr scale. The experimental data includes acceleration and VGRF data. On the one hand, the experiment compares the performance of the state of the art deep learning algorithms and the constituting components of our proposed model. On the other hand, we also evaluate the separability of the features extracted by our proposed model.

\subsubsection{Classification of Different Identities}
The performance of the learning models and the constituting components of the proposed model is reported in Table \ref{advanced_model_sdu}. Among all the multi-modal methods, DCCA achieves 96.17\% accuracy, only second to our proposed method. The results of BiLSTM and GRU are superior to the other spatial feature recognition techniques. The classification outcome of hand-crafted feature extraction is poor due to the limitations in the separation capability. Among the constituting components of the proposed model, the accuracy of the identifying features obtained by SFE and CorrMNN respectively is 95.38\% and 97.38\% respectively, lower than the performance of the hybrid model.  Our proposed method achieves 98.79\% accuracy, which shows the strong ability of the fusion model.

The ROC curve and the AUC value are presented in Fig. \ref{roc_compare_sdu}. The AUC of all the models is higher than 95\%, indicating that the gaits of 52 subjects are relatively easy to distinguish, and the performance of all the models tends to be stable. BiLSTM and DCCA are the best in the advanced models, and CorrMNN and DCMNN stand out in the constituting components.

\subsubsection{Performance of the Whole Model}

The t-SNE results are illustrated in Fig. \ref{feature_compare_sdu}. After dimensionality reduction on the original data, the boundary between different classes is not clear and the overlap is still severe. Although there are 52 classes in the dataset, the features extracted by our hybrid model are highly separable after dimensionality reduction.

\section{Conclusion}
This paper has presented a novel learning framework for distinguishing three NDDs and different severity levels of Parkinson's disease, based on human gaits and other sensory data. The constituting components in the proposed model delivers individual functionalities, e.g. the SFE model was used to extract the frequency- and time-domain features of the heterogeneous data and to encode them as Fisher vector. CorrMNN helped us to conduct feature extraction of the data and achieve proper fusion using the correlation between the signals. Multi-switch discriminator models each category of the fused features to achieve satisfactory classification. In the light of the evaluation of the performance of the three modules on the three datasets, from high to low are the hybrid model, CorrMNN, SFE and multi-switch discriminator, justifying the effectiveness of our hybrid model. Experimental results have shown that, compared with the other established methods, our proposed model resulted in better discriminative results of NDDs and severity levels of PD in an end-to-end form.


\bibliographystyle{IEEEtran}
\footnotesize
\bibliography{mybibfile}

\ifCLASSOPTIONcaptionsoff
  \newpage
\fi

\end{document}